\documentclass[journal]{IEEEtran}
\usepackage{amsmath,amsfonts}
\usepackage[algo2e,ruled,linesnumbered]{algorithm2e}
\usepackage{algorithmic}
\usepackage{algorithm}
\usepackage{array}
\usepackage[caption=false,font=normalsize,labelfont=sf,textfont=sf]{subfig}
\usepackage{textcomp}
\usepackage{stfloats}
\usepackage{url}
\usepackage{verbatim}
\usepackage{graphicx}
\usepackage{xcolor}
\usepackage[adjust]{cite}
\usepackage{graphicx}
\usepackage{rotating} 
\usepackage{multirow}
\usepackage{relsize}
\usepackage{hyperref}
\usepackage{cleveref}
\usepackage{colortbl}

\crefname{figure}{Figure}{Figures}

\SetKwFor{ForEach}{for each}{}{}
\SetKwFor{For}{for}{}{}
\SetKwFor{While}{while}{}{}
\SetKwIF{If}{ElseIf}{Else}{if}{}{else if}{else}{}

\begin{document}
\title{\LARGE\bf
An Analysis of Constraint-Based Multi-Agent Pathfinding Algorithms
}

\author{Hannah Lee$^{1}$, James D. Motes$^{1}$,  Marco Morales$^{1,2}$, and Nancy M. Amato$^{1}$% <-this % stops a space
%\thanks{*This work was not supported by any organization}% <-this % stops a space
%\thanks{Manuscript received: February 24, 2021; Revised May 26, 2021; Accepted June 23, 2021.}
%\thanks{This paper was recommended for publication by Editor M. Ani Hsieh upon evaluation of the Associate Editor and Reviewers’ comments.}
\thanks{$^{1}$Hannah Lee, James D. Motes, Marco Morales, and Nancy M. Amato are with the Parasol Lab, School of Computer Science, University of Illinois at Urbana Champaign, Champaign, IL, 61820 USA.
{\tt\small hannah9, jmotes2, moralesa, namato@illinois.edu}}%
\thanks{$^{2}$Marco Morales is also with the Department of Computer Science at Instituto Tecnol\'ogico Aut\'onomo de M\'exico (ITAM), Mexico City, M\'exico.}
%\thanks{Digital Object Identifier (DOI): see top of this page.}
}

\markboth{IEEE Transactions on Robotics}
{Lee \MakeLowercase{\textit{et al.}}: An Analysis of Constraint-Based  Multi-Agent Pathfinding  Algorithms}

% \IEEEpubidadjcol

\maketitle

\setcounter{footnote}{2}
\begin{abstract}
This study informs the design of future multi-agent pathfinding (MAPF) and multi-robot motion planning (MRMP) algorithms by guiding choices based on constraint classification for constraint-based search algorithms. 
 We categorize constraints as conservative or aggressive and provide insights into their search behavior, focusing specifically on vanilla Conflict-Based Search (CBS) and Conflict-Based Search with Priorities (CBSw/P). Under a hybrid grid-roadmap representation with varying resolution, we observe that aggressive (priority constraint) formulations tend to solve more instances as agent count or resolution increases, whereas conservative (motion constraint) formulations yield stronger solution quality when both succeed.

Findings are synthesized in a decision flowchart, aiding users in selecting suitable constraints. Recommendations extend to Multi-Robot Motion Planning (MRMP), emphasizing the importance of considering topological features alongside problem, solution, and representation features. A comprehensive exploration of the study, including raw data and map performance, is available in our public GitHub Repository\footnote{\url{https://GitHub.com/hannahjmlee/constraint-mapf-analysis}\label{git_repo}}. 
\end{abstract}


\section{Introduction}

\IEEEPARstart{M}{ulti-Agent} Pathfinding (MAPF) is a critical multi-agent coordination problem wherein safe and efficient paths are planned for multiple agents. MAPF finds applications in various domains such as assembly ~\cite{hlw-agffaptmsa-98, nb-toaraap-93, bpssk-ostaapffmarap-20}, evacuation \cite{ra-bbep-10}, formation control \cite{ba-bbfcfmt-98, tpk-ltfs-04, kh-ppfpimf-06, lsmfkk-maifice-20, lwcycl-mfifadhlatmamt-21}, localization \cite{fbkt-apptcmrl-00}, and object transportation\cite{bpssk-ostaapffmarap-20, rdjj-mfwtofar-95}.  The core challenge lies in orchestrating the concurrent traversal of agents along their respective paths without colliding with each other or with environmental obstacles. MAPF is an NP-hard problem that exhibits an exponentially growing state space with the number of agents \cite{yl-saioomrppog-13}. 

A widely adopted strategy for addressing MAPF involves using constraint-based search algorithms, which iteratively constrain the search space to identify valid solutions. Many existing algorithms, including those presented in~\cite{lb-pasfcpfwcg-11,s-cp-05,pl-ssippfde-11,dtw-paracmapfa-14}, can be generalized within this constraint-based search paradigm. In this paper, we specifically examine two prominent algorithms: Conflict-Based Search (CBS) \cite{cbs} and Conflict-Based Search with Priorities (CBSw/P) \cite{pbs}. CBS independently plans paths for individual agents, resolving conflicts by adding constraints on specific agent actions at given timesteps. CBSw/P, a variant of CBS employing prioritized planning, resolves conflicts by assigning priorities to agents, thus constraining lower-priority agents based on higher-priority ones.

\begin{figure}[!t]
\centering
\includegraphics[width=\linewidth]{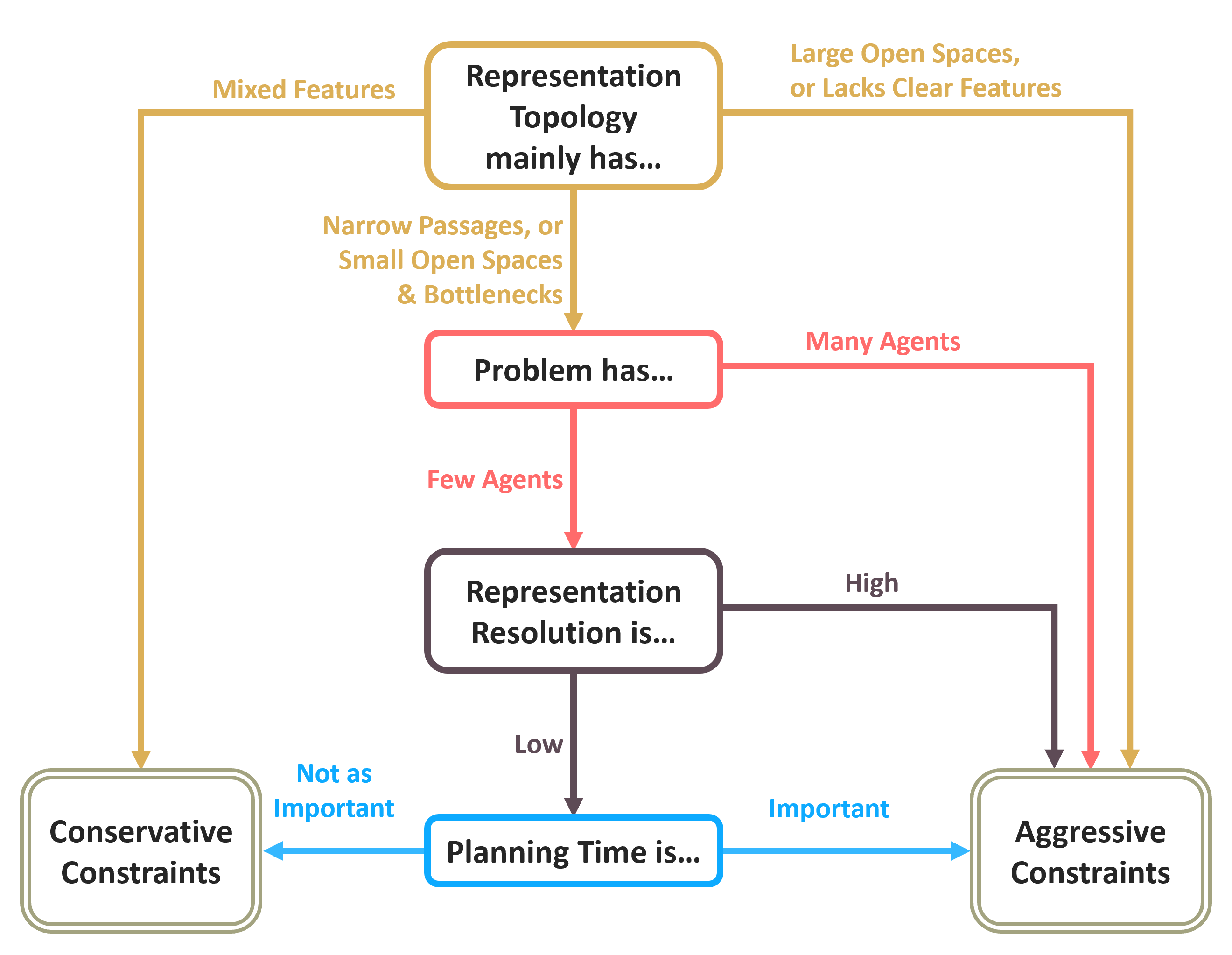}
\caption{A flowchart summarizing our findings and providing a general set of guidelines in determining when to use aggressive vs. conservative constraints given our heuristics derived from comparing vanilla CBS and CBSw/P. These decision points may shift with other search strategies or under different settings.}
\label{fig:flowchart}
\end{figure}

To effectively evaluate the performance of constraint-based search algorithms, it is essential to assess and characterize problem difficulty. Problem difficulty directly influences the efficiency of an algorithm’s conflict resolution strategy, affecting both computational complexity and solution quality. The choice of constraints significantly impacts the shape and size of the search space and is highly sensitive to problem difficulty. However, measuring the difficulty of a specific MAPF instance remains challenging due to the dynamic nature of conflict resolution in multi-agent systems. Traditional analyses estimate problem difficulty using factors such as problem size, environment topology, and environment dimensions, but these approaches do not generalize well to more complex representations, such as the roadmaps commonly used in multi-robot motion planning. This work investigates how different constraint choices affect search behavior across multiple dimensions of problem difficulty, emphasizing the importance of selecting appropriate constraints for both MAPF and more general representations.

To address this need, we introduce a framework classifying constraints as conservative (akin to CBS) or aggressive (akin to CBSw/P). Through CBS and CBSw/P, we investigate the implications of employing conservative versus aggressive constraints and analyze performance variations relative to representation topology and resolution. Our findings provide insights for informed decision-making in constraint-based search algorithm design and offer critical considerations for adapting MAPF solvers to more complex domains, such as Multi-Robot Motion Planning (MRMP).  We summarize our contributions and insights in the flowchart depicted in Figure \ref{fig:flowchart}.

\subsection{MAPF vs. MRMP}

Past studies of MAPF algorithms have concentrated mainly on the challenges presented by different environmental topologies, such as narrow passages and open areas \cite{cbs, hccbs, pbs, ssfkmwlack-mapfdvab-19}. However, these studies do not directly apply to MRMP due to a key difference in how each approaches representation. In MAPF scenarios, the emphasis is on searching within an already established representation for a team plan. In contrast, MRMP can involve generating or implicitly defining a representation that captures the robot's geometry, kinematics, and allowable motions. For instance, many sampling-based methods construct a roadmap approximating the free configuration space, whereas search-based methods can define an implicit graph through motion primitives \cite{cohen2010search}. In either case, MRMP requires more detailed modeling than MAPF. It is important to note that our paper does not explore the construction of these representations; instead, we focus on evaluating how MAPF querying techniques perform across a range of representations.

\begin{figure}[!t]
\centering
\includegraphics[width=\linewidth]{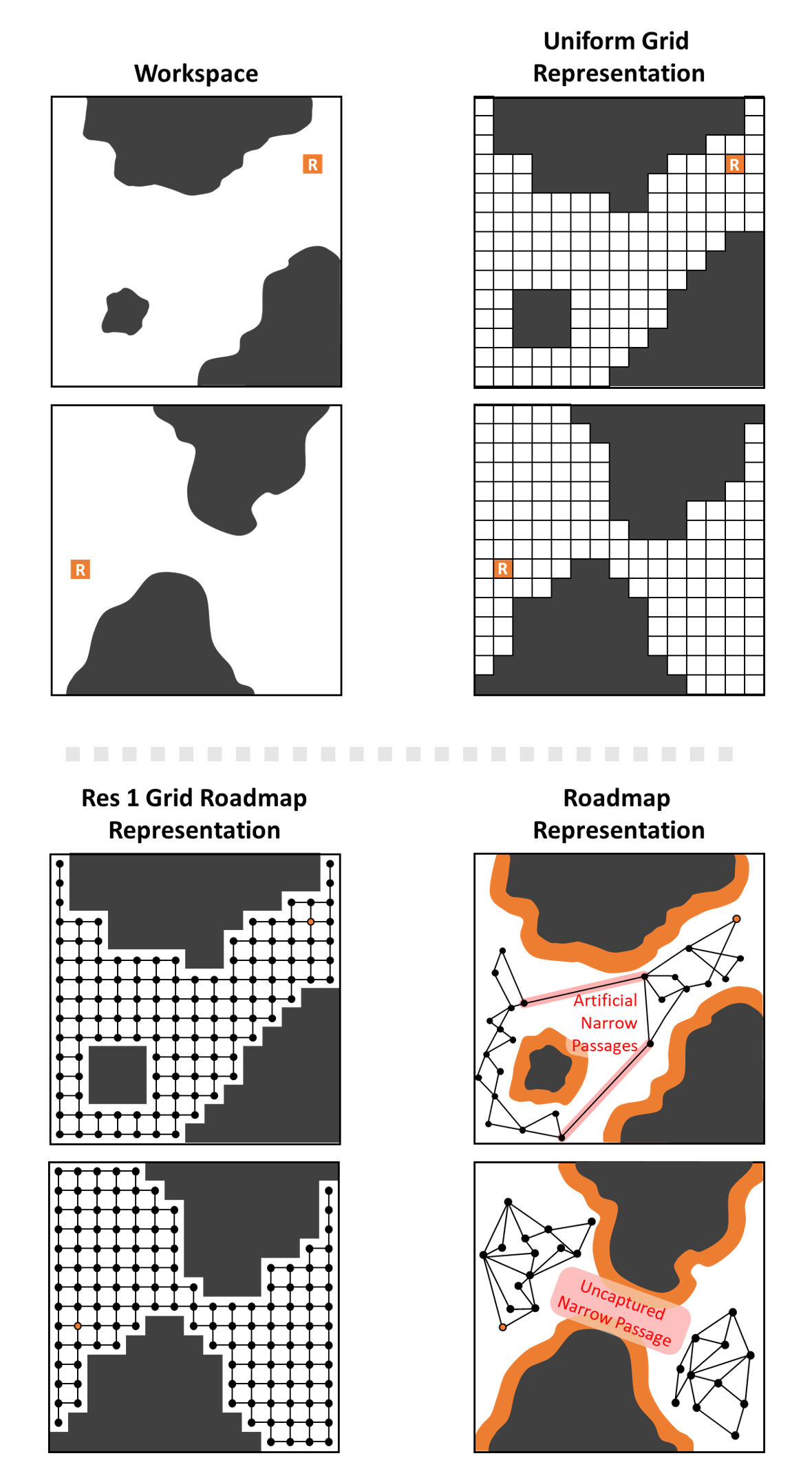}
\caption{Comparison of Grid, Grid Roadmap, and Roadmap Representations for a 2D square robot with 2 DoF. The orange position $R$ represents the robot across all three representations. In the grid representation, states are mapped to discrete cells. The grid roadmap uses a uniform grid-sampler to map states to configurations, while the roadmap representation employs a random sampler to map states to the configuration space. For the roadmap, obstacles (gray) are inflated by the robot’s radius (orange) to define the obstacle space. The top row depicts an environment without narrow passages, where random sampling inadvertently creates one. In contrast, the bottom row shows a narrow passage environment that the roadmap fails to capture, resulting in a representation that overlooks the passage. }
\label{fig:mrmp}
\end{figure}

MAPF usually coordinates the movements of several agents by mapping their states onto a uniform grid, effectively simplifying the robot's details.
On the other hand, MRMP considers the robot's shape, size, and orientation, often relying on sampling-based methods to generate non-uniform roadmaps or leveraging implicit graphs that approximate the connectivity of the free configuration space.  
Therefore, MRMP demands a more detailed representation of robots and their configuration spaces while leveraging MAPF algorithms for coordinating their motions on the generated representations. 

The standard uniform grid representation of MAPF generates a grid map that simplifies collision checking through atomic actions and edges. Thus, collision checking in MAPF is typically simple and efficient. In contrast, the non-uniform roadmap representation of MRMP is characterized by diverse edge lengths and durations. These pose collision checking challenges distinct from traditional MAPF scenarios because the robot's shape, volume, and orientation must be considered. Additionally, the variability in edge length and resolution in a non-uniform representation can give rise to artificial graph topology. 

As illustrated in the top set of images in Figure \ref{fig:mrmp},  in an environment devoid of narrow passages, sparse roadmaps with long edges can create artificial narrow passages during the search. These narrow passages, absent in the actual environment topology, emerge as a result of the representation during the search. Conversely, as illustrated in the bottom set of images in Figure \ref{fig:mrmp},  in environments with narrow passages, the representation may not capture them and result in a roadmap that lacks such passages during the search. This dynamic emphasizes the significant influence of representation topology on MRMP. The complexities arising from the varied resolutions and edge lengths underscore the challenges of directly applying MAPF methodologies to MRMP, particularly when relying solely on environmental topological analysis.

\begin{figure}[!t]
\centering
\includegraphics[width=\linewidth]{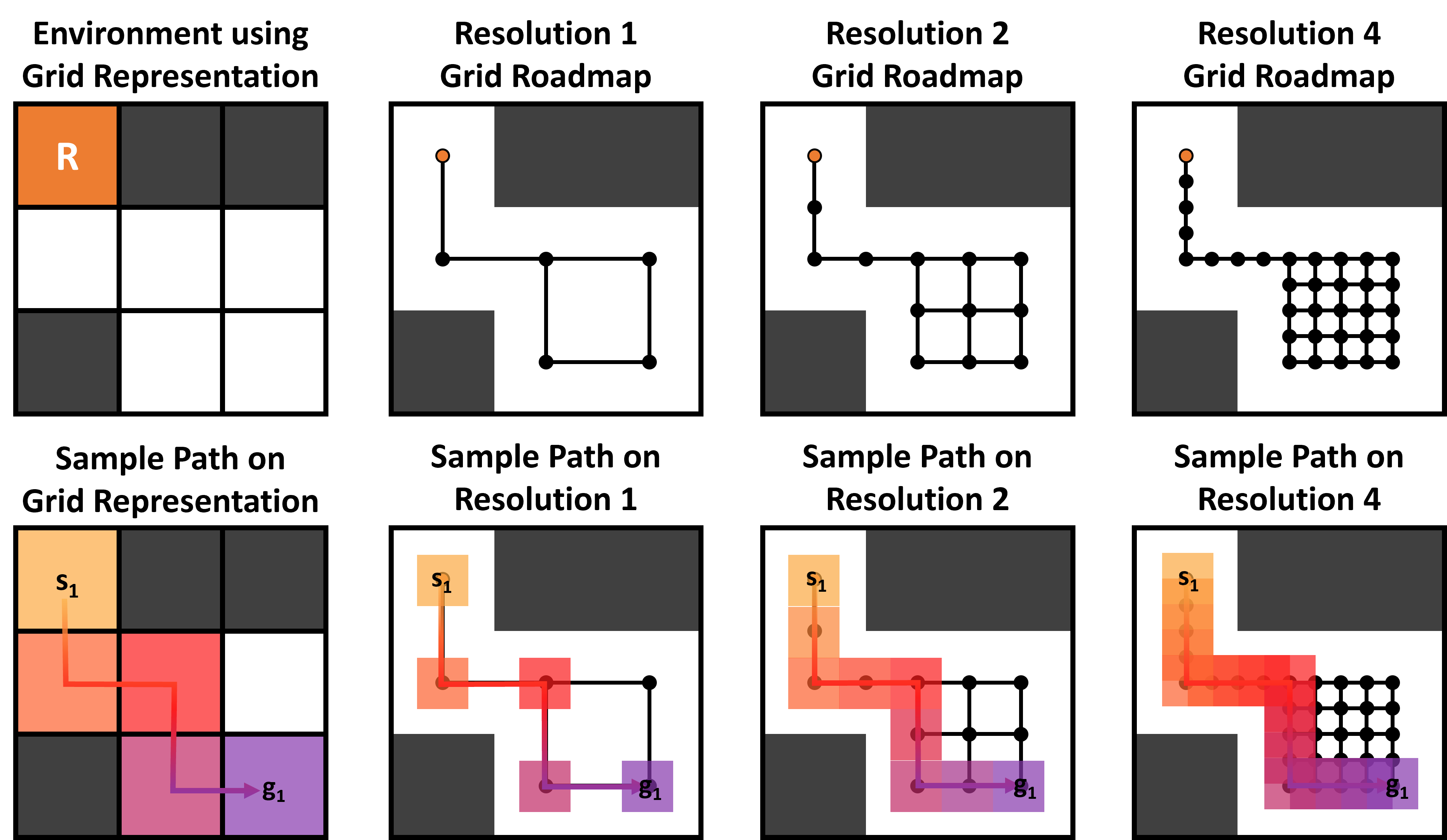}
\caption{Shown are samples of true grid and grid roadmap representations at resolutions 1, 2, and 4. This highlights cell-based planning in true grids and vertex-based planning in grid roadmaps. In grid roadmaps, resolution 1 grids equate to 4 vertices, while resolution 2 and 4 grids scale to 9 and 25 vertices, respectively. Sample paths are depicted for each resolution, showcasing increased vertex count with higher resolutions. When projecting a resolution 1 path into higher resolutions, the number of states in the path increases, akin to increasing edge discretization in a MRMP roadmap representation. }
\label{fig:resolution}
\end{figure}

In our analysis, we use a grid roadmap representation to model the MAPF problem using roadmaps. This distinction is shown in Figures \ref{fig:mrmp} and \ref{fig:resolution} and is discussed in detail in Section \ref{section:resolution}. To capture the characteristics of both MAPF grids and MRMP roadmaps, we employ a hybrid grid roadmap representation.  To clarify, our focus is not on the construction of representations; hence, we employ a simplified hybrid representation that captures  the qualities of both grid and roadmap representations.  

At a resolution of 1, the grid roadmap representation shows the same behavior as seen in traditional MAPF. As the resolution increases, we maintain the grid structure of MAPF, but mimic the behavior of denser roadmap representations encountered in MRMP. Our hybrid representation allows us to navigate the complexities of MRMP roadmap representations while retaining control over the granularity of representation resolution.  This simplified representation mitigates the possibility of an inadequate or problematic (e.g., artificial narrow passages) representation of the workspace affecting our analysis of the search process. Therefore, we operate under the assumption that we are working with a well-constructed representation, allowing us to focus solely on the comparative performance of MAPF techniques across different representations.

\subsection{Key Takeaways}

In this paper, we define aggressive and conservative constraints as a foundation for our discussion, supported by thorough experiments. We examine how these constraints affect various elements of constraint-based search algorithms. Our advice is presented in a user-friendly decision flowchart, shown in Figure \ref{fig:flowchart}, facilitating the selection of aggressive or conservative constraints for constraint-based search algorithms. Conservative constraints meticulously constrain the state space, prioritizing incremental progress towards the optimal solution while minimizing their impact during the search. In contrast, aggressive constraints aim to advance rapidly in the search space, with a broader influence that emphasizes efficient exploration. 

Deciding between aggressive and conservative constraints for a MAPF or MRMP problem involves assessing various factors, each represented by a distinct color within our flowchart: the environment representation's topology (tan), problem size (pink), representation resolution (purple), and desired planning time (blue). By systematically considering these factors, practitioners can tailor their constraint selection to align with the specific requirements and characteristics of the problem environment, ensuring optimal performance and efficient problem-solving. 

\begin{enumerate}
    \item \textit{Environment Representation Characteristics (tan):} Start by evaluating the topological features within your environment representation. If the topology consists of  predominantly large open spaces or lacks clear features, one should opt for aggressive constraints. Environments with a mix of different features should use conservative constraints. However, if comprised of small open spaces and bottlenecks or narrow passages,  move on to evaluate problem size and representation resolution. 
    \item \textit{Problem Size (pink) and Representation Resolution (purple):} For scenarios with many agents or high representation resolution, aggressive constraints can improve runtime efficiency and increase the chances of finding a solution before computational resources are exhausted. However, this comes at the cost of completeness, as more aggressive constraints may exclude valid solutions. Conversely, for smaller problem sizes and lower representation resolutions, the choice of constraints should be guided by the desired balance between solution completeness and computational performance. 
    \item \textit{Performance Expectations (blue):} Lastly, we consider the available planning time. If computational resources and runtime are not constraints, conservative constraints are a suitable choice. However, when planning time is limited, employing aggressive constraints can improve efficiency at the potential cost of completeness. 
\end{enumerate}

\subsection{Outline and Recommendations for Readers}
This work is intended to serve two primary audiences: (1) readers already familiar with constraint-based search algorithms and their applications to multi-agent systems, and (2) those seeking a deeper foundational understanding of these algorithms. 

For readers already familiar with constraint-based search algorithms, multi-agent pathfinding, and multi-robot motion planning, we recommend proceeding directly to Sections \ref{section:constraints}, \ref{section:exp_method}, and \ref{section:results}. These sections classify constraints based on their impact on search behavior, present techniques for selecting constraints, and analyze the role of representation topology, including methods for estimating problem difficulty. Sections \ref{section:constraints} and \ref{section:results} offer insights into improving algorithm design, while Section \ref{section:results} validates our findings through extensive experimental evaluation. 

For readers seeking a more comprehensive and foundational understanding, we encourage reviewing all sections, as they establish the necessary context for our discussions. Sections \ref{section:problem-def} and \ref{section:background} provide a comprehensive introduction to multi-agent pathfinding, multi-robot motion planning, and constraint-based search algorithms, ensuring a solid grasp of the foundational concepts. This work examines the search properties of constraints using the most basic version of constraint-based search algorithms to provide a clear and controlled analysis. Section \ref{section:related-work} highlights recent advancements that improve scalability for large MAPF problems and extend applications beyond MAPF, offering new practitioners a starting point for exploring constraint-based search algorithms. Section \ref{section:exp_method} details our methodology for evaluating and estimating problem difficulty, while Section \ref{section:analysis_results} presents and analyzes our experimental results, reinforcing the key claims made throughout this work.

\subsection{Scope and Limitations}

This study characterizes how constraint formulation influences behavior by comparing vanilla CBS (motion constraints) with CBSw/P (priority constraints) under the hybrid grid–roadmap discretizations evaluated in Section \ref{section:results}. Our goal is to provide controlled intuition about constraint families, not to exhaustively evaluate search strategies. Accordingly, we do not isolate or vary the high-level search or its heuristics (e.g., depth-first search PBS \cite{pbs}, bounded-suboptimal searches \cite{bssf-svotcbsftmapfp-14}, heuristic-augmented searches \cite{lrk-eecbsabssfmapf-21},  or LNS-style improvements \cite{li2021anytime}), and we do not claim that results for “conservative vs. aggressive” constraints generalize across those recent variants. The guidance offered here is most reliable for the vanilla versions and settings we study and should be treated as a first-pass heuristic when configuring newer methods. When contemporary high-level search algorithms, heuristics, or recent enhancements are employed, the observed trade-offs may shift.

In summary, our conclusions are scoped to vanilla CBS and CBSw/P using best-first searches under the representations and evaluation protocol described herein, and are not intended as general prescriptions for all conservative versus aggressive methods.

\section{Problem Definition} \label{section:problem-def}

In our discussion of the classical MAPF problem, we will be referring to the problem described in \cite{ssfkmwlack-mapfdvab-19}, albeit with some variations in notation for simplicity and to maintain consistency with prior research. It is important to note that several MAPF variants exist, each contingent on specific assumptions about the problem definition. These variants address scenarios such as distributed settings, weighted graphs, feasibility constraints, large geometric or volumetric agents, and kinematic constraints  \cite{ssfkmwlack-mapfdvab-19}. 

In the classical MAPF problem \cite{ssfkmwlack-mapfdvab-19}, we are given an input consisting of $n$ agents $a_1, a_2, \dots, a_n$. Each agent $a_i$ is characterized by a start position $s_i$ and goal position $g_i$. A path for an agent is denoted as $\pi_i$ and is defined as the sequence of actions leading from its starting position $s_i$ to its designated goal position $g_i$. The primary objective is to devise conflict-free paths for all $n$ agents within a graph $G$. 

The graph $G = (V, E)$ is an undirected graph. Vertices $V$ represent valid positions within the environment, while edges $E$ signify permissible actions with duration to transition from one vertex to an adjacent one. Agents, at each timestep, are located on a graph vertex $v \in V$. They can opt to move along an edge $(v, v') \in E$ or remain stationary at the current vertex $v$. For our grid roadmap representation, we assume a 4-neighbor movement model, where adjacent vertices are positioned directly up, down, right, and left of the current vertex.  Similar to that of a true MAPF grid representation, the grid roadmap representation has atomic edges with a duration of one timestep per edge.

The graph's representation resolution defines the granularity in the discretization of the configuration space. Higher resolution results in a more detailed and precise representation, enabling a clearer understanding of the agents' potential states and interactions. This is achieved through a finer grid, where sampling is conducted in a grid pattern with uniform spacing between points, as illustrated in Figure \ref{fig:resolution}. 

The MAPF solver aims to generate a team plan $\pi = \{\pi_1, \dots, \pi_n\}$ comprising each agent's path $\pi_i$. A path is represented as a sequence of vertices, denoted as $\pi_i = (s_i, v, \dots, v', g_i)$. The position of an agent $a_i$ at timestep $t$ is indicated as $\pi_i[t]$, corresponding to the $t$th vertex within $\pi_i$. 

\subsection{Conflicts}

In MAPF, the objective is to devise a team plan that ensures the absence of conflicts among agents. A team plan is deemed valid if no conflicts exist between any pair of agents. For the classical MAPF problem, two types of conflicts are defined: \emph{vertex conflict} and \emph{edge conflict}. These conflicts are established based on the paths of a pair of agents $a_i$ and $a_j$.  

A \emph{vertex conflict} between a pair of agent paths arises when both agents occupy the same vertex at the same time. Consequently, a vertex conflict between agents $a_i$ and $a_j$ at timestep $t$ indicates the simultaneous presence of both agents at vertex $v$. This conflict is represented using a tuple $\langle a_i, a_j,  v, t \rangle$. 

An \textit{edge conflict} arises when a pair of agents attempts to traverse the same edge simultaneously, regardless of the direction. Specifically, for agents $a_i$ and $a_j$ at timestep $t$, an edge conflict indicates their simultaneous attempt to traverse edge $(v, v')$ between timesteps $t$ and $t + 1$. This traversal can occur in either direction. When the pair of agents traverses the same edge in opposite directions, it is also known as a swapping conflict.  

 Edge conflicts are denoted using a tuple $\langle a_i, a_j, (v, v'), t\rangle$. Given that $G$ is an undirected graph with atomic edges, the order of vertices representing the edge is arbitrary; thus, $(v, v') = (v', v)$. However, in graphs with varying edge lengths or non-atomic edges, the order of vertices may become significant. In such cases, a conflict could arise when agents traverse an edge in opposite directions, but not when they follow each other. Thus, edge directionality should be considered in certain contexts. For our purposes, because our grid roadmap representation consists of atomic edges, we assume that the order of vertices does not matter. In the following sections, we will use vertex conflict notation when discussing conflicts, unless explicitly referring to edge conflicts.

In our grid roadmap representation, we adopt a simplified MRMP conflict checking scheme. Figure \ref{fig:resolution} illustrates our approach, where we assume a square robot with a width equal to half the edge length of a grid cell in a true grid representation. This square robot size enables us to emulate MAPF behavior within our hybrid representation. Unlike a MAPF grid representation, our hybrid representation considers the robot's geometry within a roadmap rather than planning across grid cells, as depicted in Figure \ref{fig:mrmp}. 

For vertex conflict checking, we verify whether any part of the robot's geometry intersects with obstacles or the occupied space of other robots at each robot's vertex position in the grid roadmap. For edge conflict checking, we examine the robot's geometry at intermediate points stationed at the center of each edge connecting two vertices for intersections with the occupied space of other robots. 

Representation resolution significantly influences conflict checking by dictating the level of detail in space discretization. In grid-based methods, a robot typically occupies a single cell, simplifying conflict checks to cell occupancy. However, MRMP introduces complexity with non-uniform roadmaps composed of vertices for robot configurations and edges for configuration sequences, making conflict checking more complex as these edges can contain hundreds of configurations. MRMP's complexity is further increased by considering robot shape and volume, making conflicts more likely not just in the same vertex but also between adjacent vertices as resolution increases. 

With our grid roadmap representation, at resolution 1, conflict checking in our grid roadmap resembles MAPF's straightforward grid-based approach. However, at higher resolutions the robot's volume may cover multiple vertices in our grid roadmap representation, and therefore requires thorough checks, similar to that seen in MRMP, for conflict-free paths. Thus, higher representation resolutions allow for a more precise modeling of robot positions while demanding a more intricate conflict checking process.

\subsection{Objective Function}
Two common objective functions used to evaluate MAPF solutions are \emph{makespan} and \emph{sum-of-costs} \cite{ssfsb-aecothomapfutmatsoco-16}. \emph{Makespan} is the number of timesteps required for all agents to reach their goal. Given a team plan $\pi = \{\pi_1, \dots, \pi_n\}$, the makespan cost of  $\pi$ is: 
\begin{align*}
    Makespan(\pi) = \max_{i = 1,  \dots,  n} |\pi_i|
\end{align*} 
\textit{Sum-of-costs}, also known as flowtime,  is the sum of timesteps across all agents to reach their goals. The sum-of-costs for a team plan $\pi$ is defined as: 
\begin{align*}
Sum\ of\ Costs(\pi) = \sum_{i =1}^{n} |\pi_i|    
\end{align*}
For the purpose of this paper, we use sum-of-costs as our objective function for all discussion and experiments. 

\section{Background} \label{section:background}
In this section, we will provide a brief overview of various MAPF solvers, introduce the concepts of aggressive and conservative constraints in the context of constraint-based search algorithms, and subsequently delve into a more detailed discussion of Conflict-Based Search (CBS) and Conflict-Based Search with Priorities (CBSw/P).  

MAPF has been extensively studied, leading to the development of various types of solvers. Reduction-based solvers transform the MAPF problem into well-known NP-hard problems like Boolean Satisfiability (SAT) and Constraint Satisfaction Problem (CSP) \cite{bs-osbafmapfwtsoco-19}. Examples of reduction-based solvers include MDD-SAT \cite{sfsb-esatmapfutsoco-16}, DPLL \cite{cs-daiomapfasst-21}, and MDD-SAT+ID \cite{ssfb-ioidisbomapfansboms-17}. These solvers utilize methods like integer linear programming or answer set programming to quickly solve the transformed NP-hard problems. 

Rule-based solvers introduce agent-movement rules and include algorithms like TASS \cite{khs-aptafnomapf-11}, push and swap \cite{lb-pasfcpfwcg-11}, parallel push and swap \cite{slb-mapfwseosap-12}, push and rotate \cite{dtw-paracmapfa-14}, and BIBOX \cite{s-anptppfmribcg-09}. While rule-based solvers guarantee fast and feasible solutions, they typically return lower-quality solutions. 

Search-based algorithms involve A*-based solvers such as HCA* and WHCA* \cite{s-cp-05}, OD and ID \cite{s-fostcpp-10}, EPEA* \cite{cgssbssh-peawsng-12}, M* \cite{wc-sefmrpp-15}, and hybrid solvers like ICTS \cite{ssgf-tictsfmapf-13}, CBS \cite{cbs}, ICBS \cite{bfsssbt-icbsfomapf-15}, ECBS \cite{bssf-svotcbsftmapfp-14}, CBSw/P and PBS \cite{pbs}, and HC-CBS \cite{hccbs}. Search-based algorithms, although slower compared to other solvers, typically yield higher-quality solutions. However, they face challenges with larger problem sizes due to the exponential state space of MAPF. For a comprehensive survey of other types and classes of MAPF solvers, please refer to \cite{fssbgsws-sbosftmapfpsac-17, s-mapfao-19, lsj-asotmapp-21}. 

Lastly, more recently, large neighborhood search (LNS) techniques for MAPF have been used to iteratively refine solutions by selectively destroying and reoptimizing parts of the plan, balancing exploration and exploitation \cite{li2021anytime}. Variants like LNS2 extend this idea by more sophisticated neighborhood selection and repair strategies, often leading to faster convergence on high-quality solutions in large or complex problem instances \cite{wang2025lns2+, li2022mapf}. While these methods typically relax completeness and optimality guarantees, their scalability and practical performance make them appealing for real-world applications.

{
\begin{algorithm2e}[!t]
\caption{Constraint-Based Search Framework}\label{alg:algorithm}
\KwData{MAPF Problem Instance}
\KwResult{Team Plan or Failure}
$CT \gets \emptyset$ 

Initialize $Root$ with low-level path for each agent

Insert $Root$ into $CT$;

\While{$CT$ \upshape{not empty}}{
$N \gets$ Lowest-cost node in $CT$

$X \gets$ Find conflict in $N$

\If{$N$ \upshape{has no conflict}}{
\Return $N.solution$
}

\ForEach {\upshape{agent} $a_i$ in $X$}{
$C \gets$ \FuncSty{ConflictResolution}($a_i$, $X$)

$A \gets N$

Add $C$ to $A.constraints$ for $a_i$

Update $A.solution$ with low-level path for $a_i$

Update $A.cost$

Insert $A$ into $CT$
}
}
\Return Failure
\end{algorithm2e}
}

\subsection{Constraint-Based Search Algorithms} \label{section:related-work}

Constraint-based search algorithms are a category of MAPF solvers that aim to find solutions for teams by progressively applying constraints to the state spaces of individual agents. These algorithms navigate the state space, prioritizing states that best optimize a specified objective function. They operate within a common framework, detailed in Algorithm \ref{alg:algorithm}. 

Constraint-based search algorithms employ both low-level and high-level searches. The low-level search involves a pathfinding algorithm that determines the sequence of actions for an agent to traverse from its start to goal position, considering a set of constraints. Conflicts are identified in a high-level search, which explores a Conflict Tree (CT). The CT iteratively resolves conflicts by growing sets of constraints. These constraints dictate the valid states during the low-level search in individual agents' state spaces. 

The CT is a binary tree. A CT Node, denoted as $N$, consists of a solution $N.solution$, a cost $N.cost$, and a set of constraints $N.constraints$. All CT nodes are consistent, meaning the paths they hold always conform to the placed constraints. The high-level search aims to find the lowest-cost node that is valid (consistent and conflict-free). 

The algorithm starts by computing independent low-level paths for each agent, creating the initial solution for the root node $Root$ of the conflict tree $CT$ (lines 1-3). It then iteratively selects the lowest-cost node $N$ from the $CT$ (line 5), identifies conflicts $X$ in the selected node (line 6), and attempts to resolve these conflicts (line 11). If no conflicts are present, the algorithm returns the solution of the current node. Otherwise, it explores alternative paths for the agents by generating constraints used to generate successor nodes (lines 12-15). The process continues until a solution is found or until the $CT$ is empty, indicating the absence of a viable solution. For a more in-depth explanation of constraint-based algorithms, please refer to \cite{fssbgsws-sbosftmapfpsac-17}. 

Constraint-based search algorithms typically employ two main approaches: modifying constraints or adjusting the search process. In some variants, constraint refinement focuses on smarter conflict resolution strategies, such as introducing more refined constraints that prevent future conflicts with minimal impact on the rest of the plan. These modifications emphasize conflict resolution and constraint propagation \cite{cbs, pbs, lhsfmk-dsfmapfwcbs-19, lhsmk-sbcfgbmapf-19, lghsmk-ntfpsbimapf-20}. Conflict resolution strategies play a crucial role in guiding the search, ensuring efficient exploration of the state space. The algorithm's effectiveness lies in its ability to balance exploration and exploitation, navigating through the individual state spaces of the agents to find a feasible solution.

Alternatively, search modification algorithms enhance the efficiency of the high-level or low-level search through techniques like heuristic-driven pruning or parallel processing. These methods aim to improve the performance of constraint-based search by incorporating advanced search heuristics and allowing for controlled suboptimality \cite{hccbs, bfsssbt-icbsfomapf-15, bfhsclk-idcbs-20, clbmckk-ahtcbsfmapf-18, bssf-svotcbsftmapfp-14, lrk-eecbsabssfmapf-21, lfbmk-ihfmapfwcbs-19, ssfs-macbsfomapf-12, svsab-omapf-19}. Even when algorithms utilize both approaches, their contributions can generally be categorized as either modifications to the constraints or modifications to the search process \cite{l-mapffla-19, o-lsbafqmapf-23, aysas-mapfwct-22, jzzjyh-cbswdlafrppiude-23, a-mapfwkcvcbs-20}.

This study focuses on modifications to constraints rather than alterations to the search process. Although constraint-based search algorithms share a common framework, their performance and behavior vary depending on how conflicts are resolved through the application of constraints to the state space. Our approach emphasizes selecting and refining constraints based on key problem features, such as problem size, representation resolution, and representation topology. We do not explore modifications to the search process, such as heuristic-driven improvements, as these are outside the scope of this paper. By tailoring constraints to specific problem characteristics, we aim to enhance the design and performance of constraint-based search algorithms across diverse domains. We categorize constraints as either aggressive or conservative and, in the following subsections, define these categorizations and discuss the conflict resolution techniques of CBS and CBSw/P.

\subsection{Aggressive vs. Conservative Constraints} \label{section:constraints}

\textit{Conservative constraints} prioritize maintaining a complete search space and are characterized by their localized impact on agent movements. These constraints impose precise, highly specific limitations on agents, aiming to resolve conflicts with minimal disruption to their overall trajectories. They prioritize precision in agent paths by localizing the impact of their constraints. Typically, the precision and limited scope of these constraints result in solutions with more coordinated behavior. However, this approach can lead to longer planning times when numerous constraints are required to reach a solution, as observed in our experiments (Section \ref{section:analysis_results}). 

\textit{Aggressive constraints}, on the other hand, prioritize rapid advancement in the search space by constraining a broader range of states, often influencing significant portions of the agent's trajectory. While these constraints may be applied across the entire path, their primary aim is to encompass more states to expedite solution discovery. This emphasis on efficient exploration typically leads to faster planning but may compromise path quality. 
The broad nature of aggressive constraints typically results in solutions with less coordinated behavior. 

Aggressive constraints encourage exploratory behavior, enhancing search breadth, whereas conservative constraints focus on exploitation, efficiently resolving individual conflicts and maintaining solution completeness. Choosing between these strategies requires weighing path quality against planning efficiency within the given context.

We use conservative to mean motion constraints as in vanilla CBS, and aggressive to mean priority constraints as in CBSw/P. These terms describe how conflicts are turned into constraints. They are independent of the high-level search strategy (e.g., best-first, depth-first, weighted, or LNS). Thus, our comparisons isolate the effects of constraint families under a fixed high-level search, rather than conflating them with changes in the search itself.

\subsection{Conflict-Based Search}

In Conflict-Based Search (CBS), when a conflict $\langle a_i, a_j, v, t \rangle$ is detected between agents $a_i$ and $a_j$ at vertex $v$ during timestep $t$, its conflict resolution involves creating two constraints: $\langle a_i, v, t \rangle$ and $\langle a_j, v, t \rangle$. These constraints, termed \textit{motion constraints}, specify that a particular agent is restricted from accessing a vertex or edge at a specific timestep. CBS ensures optimality and a complete search space by generating two constraints that independently limit the low-level searches for agents $a_i$ and $a_j$. Exploring both possibilities guarantees that no states within the search space are overlooked or excluded. 

The motion constraints utilized by CBS are classified as conservative constraints. They pinpoint a specific position at a precise moment, minimizing the number of constrained states. This approach ensures the preservation of a complete search space, a hallmark of conservative strategies. Unlike coupled approaches that search joint composite state spaces to maintain completeness, CBS offers distinct advantages by searching the individual state spaces of each agent. Consequently, CBS's state space scales linearly with the number of robots, while its search space expands with the number of conflicts encountered. As a result, CBS's performance is influenced by the problem's complexity, the environment's topology, and the frequency of conflicts during the solving process.  Each conflict encountered results in CBS adding two additional child nodes to its CT, leading to increases in the CT size, especially in environments with high-traffic areas. 

\subsection{Conflict-Based Search with Priorities}

In Conflict-Based Search with Priorities (CBSw/P), when a conflict $\langle a_i, a_j, v, t \rangle$ is detected, its conflict resolution involves creating two \textit{priority constraints}: $\langle a_i \prec a_j \rangle$ and $\langle a_j \prec a_i \rangle$.  The first constraint denotes that agent $a_i$ has a higher priority than agent $a_j$. When agent $a_i$ possesses higher priority than agent $a_j$, the path for agent $a_i$ is planned first. Then, the path for agent $a_j$ is planned with respect to agent $a_i$'s path such that agent $a_j$ completely avoids agent $a_i$'s path. This ensures that after replanning both agents, no conflicts exist between the two agents. 

A popular variant of CBSw/P is Priority-Based Search (PBS), which uses the constraint method of CBSw/P alongside a depth-first high-level search. By using the best-first high-level search of CBSw/P, we avoid conflating constraint classifications with search strategies. 

Priority constraints, classified as aggressive, require an agent to avoid any intersections with paths of higher-priority agents for their entire journey. Unlike motion constraints that apply to specific movements and timesteps, priority constraints are persistent, preventing an agent from entering states occupied by those with higher priority. These constraints might shift with every planning iteration, reflecting changes in the higher-priority agents' paths. Due to its aggressive constraints, CBSw/P is an incomplete solver that does not maintain a complete search space. 

CBSw/P's efficiency is influenced by its Priority Tree (PT) size, which is analogous to CBS's CT but employs priority constraints instead of motion constraints for expansion. The PT is often smaller than the CT due to a bounded number of branches limited by the number of all possible ordered priority pairs. Although the PT has fewer branches, CBSw/P's low-level search typically demands more time due to potentially having to solve multiple pathfinding problems at each node if priority constraints are interconnected, coupled with the higher cost of verifying constraint consistency. CBSw/P is both incomplete and suboptimal; it may fail to find a solution even when one exists and does not guarantee the lowest-cost solution, as it relies on heuristic prioritization rather than exhaustive exploration.

\subsection{CBS vs. CBSw/P}
While existing research explores the map topologies where CBS and CBSw/P excel, there is a gap in analyzing how their performance is affected by the representation topology and resolution. This absence of analysis poses challenges in extending these findings to the domain of multi-robot motion planning, where representations are typically denser, more intricate, and contain artificial topologies. Our focus on studying CBS and CBSw/P stems from their shared algorithmic structure, utilizing the hierarchical planning approach introduced by CBS, which divides the search into two levels. The primary distinction between these algorithms lies in their conflict resolution and how they constrain agent state spaces. In what follows, we compare motion vs. priority constraints while keeping the high-level search strategy the same (best-first). Analyses that vary the strategy (e.g., DFS, weighted, LNS) are left for future work.

Both CBS and CBSw/P iteratively expand their search spaces by traversing their respective high-level trees. As constraints are imposed, the search space expands accordingly, and the explored states resulting from this expansion are determined by the constraints imposed by the algorithm on their agents. 

CBS employs motion constraints to conservatively restrict an agent from accessing a specific configuration at a specific timestep. These constraints promote more coordinated behavior between agents but result in longer planning times due to the increased size of the CT and the incremental nature of constraint application, which restricts the search space one state at a time. 

In contrast, CBSw/P utilizes priority constraints, preventing an agent from accessing configurations traversed by higher-priority agents for the entire planning duration. This approach ensures that two agents conflict at most once within a given branch of the priority tree (PT), reducing the conflict-driven search expansion observed in CBS. By aggressively imposing broad path constraints, CBSw/P prioritizes efficient solution discovery over path quality. However, this aggressive constraint application leads to less coordinated behavior between agents.

\subsection{Recent Advancements}

\begin{table*}[!t]
\small
\centering
\caption{Comparison of Related Work}
{
\begin{tabular}{lccccccc}
\multicolumn{8}{l}{\textbf{MAPF (Multi-Agent Pathfinding)}} \\ \hline
\multicolumn{1}{|c|}{\textbf{Algorithm}} & \multicolumn{1}{c|}{\textbf{Constraint Type}} & \multicolumn{1}{c|}{\textbf{Complete}} & \multicolumn{1}{c|}{\textbf{Optimal}} & \multicolumn{1}{c|}{\textbf{Constraint}} & \multicolumn{1}{c|}{\textbf{Search}} & \multicolumn{1}{c|}{\textbf{Motion}} & \multicolumn{1}{c|}{\textbf{Priority}} \\ \hline
\multicolumn{1}{|l|}{CBS \cite{cbs}} & \multicolumn{1}{c|}{C} & \multicolumn{1}{c|}{$\times$} & \multicolumn{1}{c|}{$\times$} & \multicolumn{1}{c|}{$\times$} & \multicolumn{1}{c|}{} & \multicolumn{1}{c|}{$\times$} & \multicolumn{1}{c|}{} \\ \hline
\multicolumn{1}{|l|}{CBSw/P \cite{pbs}} & \multicolumn{1}{c|}{A} & \multicolumn{1}{c|}{} & \multicolumn{1}{c|}{} & \multicolumn{1}{c|}{$\times$} & \multicolumn{1}{c|}{} & \multicolumn{1}{c|}{} & \multicolumn{1}{c|}{$\times$} \\ \hline
\multicolumn{1}{|l|}{PBS \cite{pbs}} & \multicolumn{1}{c|}{A} & \multicolumn{1}{c|}{} & \multicolumn{1}{c|}{} & \multicolumn{1}{c|}{} & \multicolumn{1}{c|}{$\times$} & \multicolumn{1}{c|}{} & \multicolumn{1}{c|}{$\times$} \\ \hline
\multicolumn{1}{|l|}{CBS w/ BP \cite{boyarski2015don}} & \multicolumn{1}{c|}{C} & \multicolumn{1}{c|}{$\times$} & \multicolumn{1}{c|}{$\times$} & \multicolumn{1}{c|}{} & \multicolumn{1}{c|}{$\times$} & \multicolumn{1}{c|}{$\times$} & \multicolumn{1}{c|}{} \\ \hline
\multicolumn{1}{|l|}{\begin{tabular}[c]{@{}l@{}}CBS w/ Disjoint\\ Splitting \cite{lhsfmk-dsfmapfwcbs-19}\end{tabular}} & \multicolumn{1}{c|}{C} & \multicolumn{1}{c|}{$\times$} & \multicolumn{1}{c|}{$\times$} & \multicolumn{1}{c|}{$\times$} & \multicolumn{1}{c|}{} & \multicolumn{1}{c|}{$\sim$} & \multicolumn{1}{c|}{} \\ \hline
\multicolumn{1}{|l|}{CBS w/ PC \cite{bfsssbt-icbsfomapf-15}} & \multicolumn{1}{c|}{C} & \multicolumn{1}{c|}{$\times$} & \multicolumn{1}{c|}{$\times$} & \multicolumn{1}{c|}{} & \multicolumn{1}{c|}{$\times$} & \multicolumn{1}{c|}{$\times$} & \multicolumn{1}{c|}{} \\ \hline
\multicolumn{1}{|l|}{CBSH \cite{clbmckk-ahtcbsfmapf-18}} & \multicolumn{1}{c|}{C} & \multicolumn{1}{c|}{$\times$} & \multicolumn{1}{c|}{$\times$} & \multicolumn{1}{c|}{} & \multicolumn{1}{c|}{$\times$} & \multicolumn{1}{c|}{} & \multicolumn{1}{c|}{} \\ \hline
\multicolumn{1}{|l|}{CBSH-RCT \cite{lghsmk-ntfpsbimapf-20}} & \multicolumn{1}{c|}{C} & \multicolumn{1}{c|}{$\times$} & \multicolumn{1}{c|}{$\times$} & \multicolumn{1}{c|}{$\times$} & \multicolumn{1}{c|}{} & \multicolumn{1}{c|}{$\sim$} & \multicolumn{1}{c|}{} \\ \hline
\multicolumn{1}{|l|}{CBS-M \cite{zhang2022multi}} & \multicolumn{1}{c|}{C} & \multicolumn{1}{c|}{$\times$} & \multicolumn{1}{c|}{$\times$} & \multicolumn{1}{c|}{$\times$} & \multicolumn{1}{c|}{} & \multicolumn{1}{c|}{$\sim$} & \multicolumn{1}{c|}{} \\ \hline
\multicolumn{1}{|l|}{ECBS \cite{bssf-svotcbsftmapfp-14}} & \multicolumn{1}{c|}{C} & \multicolumn{1}{c|}{$\times$} & \multicolumn{1}{c|}{$\mathcal{B}$} & \multicolumn{1}{c|}{} & \multicolumn{1}{c|}{$\times$} & \multicolumn{1}{c|}{$\times$} & \multicolumn{1}{c|}{} \\ \hline
\multicolumn{1}{|l|}{EECBS \cite{lrk-eecbsabssfmapf-21}} & \multicolumn{1}{c|}{C} & \multicolumn{1}{c|}{$\times$} & \multicolumn{1}{c|}{$\mathcal{B}$} & \multicolumn{1}{c|}{} & \multicolumn{1}{c|}{$\times$} & \multicolumn{1}{c|}{$\times$} & \multicolumn{1}{c|}{} \\ \hline
\multicolumn{1}{|l|}{GPBS \cite{chan2023greedy}} & \multicolumn{1}{c|}{A} & \multicolumn{1}{c|}{} & \multicolumn{1}{c|}{} & \multicolumn{1}{c|}{} & \multicolumn{1}{c|}{$\times$} & \multicolumn{1}{c|}{} & \multicolumn{1}{c|}{$\times$} \\ \hline
\multicolumn{1}{|l|}{HC-CBS \cite{hccbs}} & \multicolumn{1}{c|}{C} & \multicolumn{1}{c|}{$\times$} & \multicolumn{1}{c|}{$\times$} & \multicolumn{1}{c|}{} & \multicolumn{1}{c|}{$\times$} & \multicolumn{1}{c|}{$\times$} & \multicolumn{1}{c|}{} \\ \hline
\multicolumn{1}{|l|}{IDCBS \cite{bfhsclk-idcbs-20}} & \multicolumn{1}{c|}{C} & \multicolumn{1}{c|}{$\times$} & \multicolumn{1}{c|}{$\times$} & \multicolumn{1}{c|}{} & \multicolumn{1}{c|}{$\times$} & \multicolumn{1}{c|}{$\times$} & \multicolumn{1}{c|}{} \\ \hline
\multicolumn{1}{|l|}{LaCAM \cite{o-lsbafqmapf-23}} & \multicolumn{1}{c|}{C} & \multicolumn{1}{c|}{$\times$} & \multicolumn{1}{c|}{$\times$} & \multicolumn{1}{c|}{$\times$} & \multicolumn{1}{c|}{$\times$} & \multicolumn{1}{c|}{$\times$} & \multicolumn{1}{c|}{} \\ \hline
\multicolumn{1}{|l|}{Lazy CBS \cite{gange2019lazy}} & \multicolumn{1}{c|}{C} & \multicolumn{1}{c|}{$\times$} & \multicolumn{1}{c|}{$\times$} & \multicolumn{1}{c|}{} & \multicolumn{1}{c|}{$\times$} & \multicolumn{1}{c|}{$\times$} & \multicolumn{1}{c|}{} \\ \hline
\multicolumn{1}{|l|}{MA-CBS \cite{ssfs-macbsfomapf-12}} & \multicolumn{1}{c|}{C} & \multicolumn{1}{c|}{$\times$} & \multicolumn{1}{c|}{$\times$} & \multicolumn{1}{c|}{} & \multicolumn{1}{c|}{$\times$} & \multicolumn{1}{c|}{$\times$} & \multicolumn{1}{c|}{} \\ \hline
 & \multicolumn{1}{l}{} & \multicolumn{1}{l}{} & \multicolumn{1}{l}{} & \multicolumn{1}{l}{} & \multicolumn{1}{l}{} & \multicolumn{1}{l}{} & \multicolumn{1}{l}{} \\
\multicolumn{8}{l}{\textbf{MAPD (Multi-Agent Pickup and Delivery)}} \\ \hline
\multicolumn{1}{|c|}{\textbf{Algorithm}} & \multicolumn{1}{c|}{\textbf{Constraint Type}} & \multicolumn{1}{c|}{\textbf{Complete}} & \multicolumn{1}{c|}{\textbf{Optimal}} & \multicolumn{1}{c|}{\textbf{Constraint}} & \multicolumn{1}{c|}{\textbf{Search}} & \multicolumn{1}{c|}{\textbf{Motion}} & \multicolumn{1}{c|}{\textbf{Priority}} \\ \hline
\multicolumn{1}{|l|}{LNS-PBS \cite{xu2022multi}} & \multicolumn{1}{c|}{A} & \multicolumn{1}{c|}{$\times$} & \multicolumn{1}{c|}{} & \multicolumn{1}{c|}{} & \multicolumn{1}{c|}{$\times$} & \multicolumn{1}{c|}{} & \multicolumn{1}{c|}{$\times$} \\ \hline
\multicolumn{1}{|l|}{LNS-wPBS \cite{xu2022multi}} & \multicolumn{1}{c|}{A} & \multicolumn{1}{c|}{} & \multicolumn{1}{c|}{} & \multicolumn{1}{c|}{} & \multicolumn{1}{c|}{$\times$} & \multicolumn{1}{c|}{} & \multicolumn{1}{c|}{$\times$} \\ \hline
 & \multicolumn{1}{l}{} & \multicolumn{1}{l}{} & \multicolumn{1}{l}{} & \multicolumn{1}{l}{} & \multicolumn{1}{l}{} & \multicolumn{1}{l}{} & \multicolumn{1}{l}{} \\
\multicolumn{8}{l}{\textbf{MAPF-CT (Multi-Agent Pathfinding with Continuous Time)}} \\ \hline
\multicolumn{1}{|c|}{\textbf{Algorithm}} & \multicolumn{1}{c|}{\textbf{Constraint Type}} & \multicolumn{1}{c|}{\textbf{Complete}} & \multicolumn{1}{c|}{\textbf{Optimal}} & \multicolumn{1}{c|}{\textbf{Constraint}} & \multicolumn{1}{c|}{\textbf{Search}} & \multicolumn{1}{c|}{\textbf{Motion}} & \multicolumn{1}{c|}{\textbf{Priority}} \\ \hline
\multicolumn{1}{|l|}{CBICS \cite{walker2021conflict}} & \multicolumn{1}{c|}{C} & \multicolumn{1}{c|}{$\times$} & \multicolumn{1}{c|}{$\times$} & \multicolumn{1}{c|}{$\times$} & \multicolumn{1}{c|}{} & \multicolumn{1}{c|}{$\sim$} & \multicolumn{1}{c|}{} \\ \hline
\multicolumn{1}{|l|}{CCBS \cite{andreychuk2022multi}} & \multicolumn{1}{c|}{C} & \multicolumn{1}{c|}{$\times$} & \multicolumn{1}{c|}{$\times$} & \multicolumn{1}{c|}{$\times$} & \multicolumn{1}{c|}{} & \multicolumn{1}{c|}{$\sim$} & \multicolumn{1}{c|}{} \\ \hline
 & \multicolumn{1}{l}{} & \multicolumn{1}{l}{} & \multicolumn{1}{l}{} & \multicolumn{1}{l}{} & \multicolumn{1}{l}{} & \multicolumn{1}{l}{} & \multicolumn{1}{l}{} \\
\multicolumn{8}{l}{\textbf{MRMP (Multi-Robot Motion Planning)}} \\ \hline
\multicolumn{1}{|c|}{\textbf{Algorithm}} & \multicolumn{1}{c|}{\textbf{Constraint Type}} & \multicolumn{1}{c|}{\textbf{Complete}} & \multicolumn{1}{c|}{\textbf{Optimal}} & \multicolumn{1}{c|}{\textbf{Constraint}} & \multicolumn{1}{c|}{\textbf{Search}} & \multicolumn{1}{c|}{\textbf{Motion}} & \multicolumn{1}{c|}{\textbf{Priority}} \\ \hline
\multicolumn{1}{|l|}{CBS-MP \cite{solis2021representation}} & \multicolumn{1}{c|}{C} & \multicolumn{1}{c|}{$\times$} & \multicolumn{1}{c|}{$\times$} & \multicolumn{1}{c|}{$\times$} & \multicolumn{1}{c|}{} & \multicolumn{1}{c|}{$\sim$} & \multicolumn{1}{c|}{} \\ \hline
\multicolumn{1}{|l|}{D-PBS \cite{zhang2024d}} & \multicolumn{1}{c|}{A} & \multicolumn{1}{c|}{} & \multicolumn{1}{c|}{} & \multicolumn{1}{c|}{$\times$} & \multicolumn{1}{c|}{} & \multicolumn{1}{c|}{} & \multicolumn{1}{c|}{$\times$} \\ \hline
\multicolumn{1}{|l|}{S2M2 \cite{chen2021scalable}} & \multicolumn{1}{c|}{A} & \multicolumn{1}{c|}{} & \multicolumn{1}{c|}{} & \multicolumn{1}{c|}{} & \multicolumn{1}{c|}{$\times$} & \multicolumn{1}{c|}{} & \multicolumn{1}{c|}{$\times$} \\ \hline
\end{tabular}%
}
\label{table:related-work}
\end{table*}

In this study, we employ the vanilla versions of Conflict-Based Search (CBS) and Priority-Based Search (CBSw/P) to highlight how different constraint mechanisms impact search efficiency. Despite numerous enhancements in the literature, most modern MAPF approaches still build on CBS’s motion constraints or CBSw/P’s priority constraints. The choice between these paradigms significantly affects scalability, solution quality, and runtime, emphasizing the importance of constraint selection for specific problem domains. However, baseline CBS often struggles with scalability as the number of agents increases, leading to various extensions that modify the search procedure, refine constraints, or balance optimality with runtime. Table \ref{table:related-work} categorizes constraint-based search algorithms based on (1) aggressive (A) vs. conservative (C) constraints, (2) completeness, (3) optimality or bounded suboptimality ($\mathcal{B})$, (4) whether they modify the search or the constraints, and (5) whether they use CBS’s motion constraints or CBSw/P’s priority constraints. Below, we discuss some key methods in more detail.

CBS is inherently conservative, introducing collision-avoidance constraints only when conflicts occur, minimizing disruption to other agents and timesteps. This localized conflict resolution ensures completeness (it always finds a solution if one exists) and optimality (it guarantees a minimal-cost solution given sufficient time). Several extensions build on this foundation:
\begin{itemize}
    \item CBS with Disjoint Splitting \cite{lhsfmk-dsfmapfwcbs-19} maintains CBS’s conservative approach but reduces branching overhead by introducing positive constraints instead of two negative constraints. This improves CBS’s success rate and runtime by up to two orders of magnitude. 
    \item CBS with Heuristics (CBSH) \cite{clbmckk-ahtcbsfmapf-18} and CBSH-RCT \cite{lghsmk-ntfpsbimapf-20} enhance the high-level search using heuristics. CBSH assigns additional costs to conflict tree nodes based on conflict types, improving success rate and runtime by up to 50×. CBSH-RCT further accelerates resolution by applying constraints over extended time windows to target symmetry conflicts, doubling success rates and improving runtime by an order of magnitude. Both maintain completeness and optimality. 
    \item Meta-Agent CBS (MA-CBS) \cite{ssfs-macbsfomapf-12} merges frequently conflicting agents into meta-agents, handling them in a composite state during low-level planning. This approach improves CBS’s success rate by a factor of two and enhances runtime efficiency.     
    \item CBS with Bypass (CBS w/ BP) \cite{boyarski2015don} reduces conflict tree expansions by resolving conflicts through alternative, cost-equivalent paths with fewer conflicts, avoiding unnecessary node splitting. This optimization maintains CBS’s completeness and optimality guarantees while improving efficiency.
    \item CBS with Prioritized Conflicts (CBS w/ PC) \cite{bfsssbt-icbsfomapf-15} classifies conflicts as cardinal, semi-cardinal, or non-cardinal based on their impact on solution cost. By prioritizing more problematic conflicts, CBS w/ PC guides the high-level search toward an optimal solution more efficiently while preserving CBS’s completeness and optimality.
    \item Hierarchical Composition CBS (HC-CBS) \cite{hccbs} and its parallel variants improve CBS by resolving conflicts hierarchically, reducing computational complexity through multi-level abstraction. Parallel HC-CBS (PHC-CBS) and Dynamic Parallel HC-CBS (DPHC-CBS) further enhance efficiency by distributing computations across multiple threads to accelerate runtime.
\end{itemize}
All these methods retain CBS’s conservative constraint style by keeping constraints localized in time and space, ensuring optimal solutions while improving efficiency.

To improve scalability while relaxing strict optimality, several CBS extensions retain the conservative conflict model but introduce modifications that prioritize efficiency over exact solution quality:
\begin{itemize}
    \item Enhanced CBS (ECBS) \cite{bssf-svotcbsftmapfp-14} and Enhanced Extended CBS (EECBS) \cite{lrk-eecbsabssfmapf-21} introduce a suboptimality bound $w>1$, maintaining CBS’s localized, conflict-triggered constraints while employing aggressive search pruning through inflated heuristics or priority queues. These methods significantly reduce runtime while ensuring solutions remain within a bounded factor of optimal.
    \item Lazy CBS \cite{gange2019lazy} formulates constraints dynamically rather than predefining all constraints and variables upfront. Instead of branching on every detected conflict, it incrementally adds violated constraints to a partial constraint model, using core-guided search to find feasible minimum-cost plans. This approach improves robustness and runtime, particularly in highly contentious scenarios.
\end{itemize}
Although these methods improve planning efficiency by pruning or restructuring search strategies, they retain CBS’s conservative constraint formulation. Their primary distinction from vanilla CBS lies in how they explore, expand, and manage the search tree rather than in how they define constraints.

In contrast to CBS, Conflict-Based Search with Priorities (CBSw/P) applies a more aggressive constraint strategy. Agents are assigned a fixed priority ordering, and when a conflict arises, the lower-priority agent must replan its entire trajectory to avoid conflicts with higher-priority agents for the full time horizon. As a result, even a single conflict can lead to broad constraints that reshape an agent's overall path. Priority-Based Search (PBS) then applies a depth-first high-level search instead of a best-first level search. Several PBS-based extensions build on this approach:
\begin{itemize}
    \item LNS-PBS and LNS-wPBS \cite{xu2022multi} integrate large neighborhood search (LNS) with PBS, iteratively modifying and re-optimizing subsets of agent paths. LNS-PBS maintains completeness for well-formed problem instances, focusing on effectiveness and solution quality. LNS-wPBS, however, incorporates a windowed MAPF algorithm that sacrifices completeness guarantees but significantly enhances scalability and stability.
    \item Greedy PBS (GPBS) \cite{chan2023greedy} employs a greedy strategy to establish or update agent priorities, aiming to minimize collisions while improving efficiency. It introduces techniques such as partial expansions, target reasoning, induced constraints, and soft restarts to further optimize search performance. GPBS demonstrates higher success rates compared to other suboptimal algorithms, particularly in MAPF instances with small environments or dense obstacles.
\end{itemize}
These approaches leverage PBS’s aggressive constraint application to improve efficiency, albeit often at the cost of completeness.

Beyond the core CBS and CBSw/P families, several methods extend MAPF to specialized problem variations (see Table \ref{table:related-work}):
\begin{itemize}
    \item Multi-Agent Pickup and Delivery (MAPD) methods, including certain CBSw/P extensions, introduce dynamic task arrivals and often prioritize responsiveness over completeness or optimality.
    \item MAPF with Continuous Time (MAPF-CT) generalizes CBS’s local constraints to continuous time intervals rather than discrete timesteps. These methods maintain CBS’s conservative constraint application, preserving completeness and, in many cases, optimality.
    \item Multi-Robot Motion Planning (MRMP) extends MAPF to higher-dimensional spaces with complex kinematic or dynamic constraints. Many MRMP approaches adapt conflict- or priority-based frameworks to continuous spaces, balancing completeness and computational efficiency.
\end{itemize}

While CBS and CBSw/P represent the dominant paradigms for motion- vs. priority-based constraints, some methods adopt a more unified constraint-based approach. One such example is Lazy Constraint Addition for MAPF (LaCAM) \cite{o-lsbafqmapf-23}, which encodes all agent movements, collision avoidance, and cost objectives into a single constraint model. Unlike CBS, which branches based on conflicts, or CBSw/P, which assigns global priorities, LaCAM lazily adds constraints only when a partial solution exhibits a collision. This method retains completeness when exhaustively implemented but produces suboptimal solutions. By integrating advanced constraint-solving techniques like clause learning with MAPF-specific collision detection, LaCAM achieves high success rates and low runtimes, particularly in large-agent scenarios.

Each algorithm finds its niche by balancing aggressiveness vs. conservativeness in constraint application, completeness vs. bounded suboptimality, and reliance on explicit high-level/low-level expansions vs. unified constraint models. CBS-based methods remain the gold standard for complete and optimal planning, while CBSw/P-inspired approaches trade completeness for speed and scalability, making them well-suited for large, complex environments.

\section{Experimental Methodology} \label{section:exp_method}

In the following subsections, we discuss our choices in representation resolution, the identification of representation topology, and our experimental setup. Together, these aspects provide a comprehensive analysis of the behavior of aggressive and conservative constraints under varying problem properties.

\begin{figure}[!t]
\centering
\includegraphics[width=\linewidth]{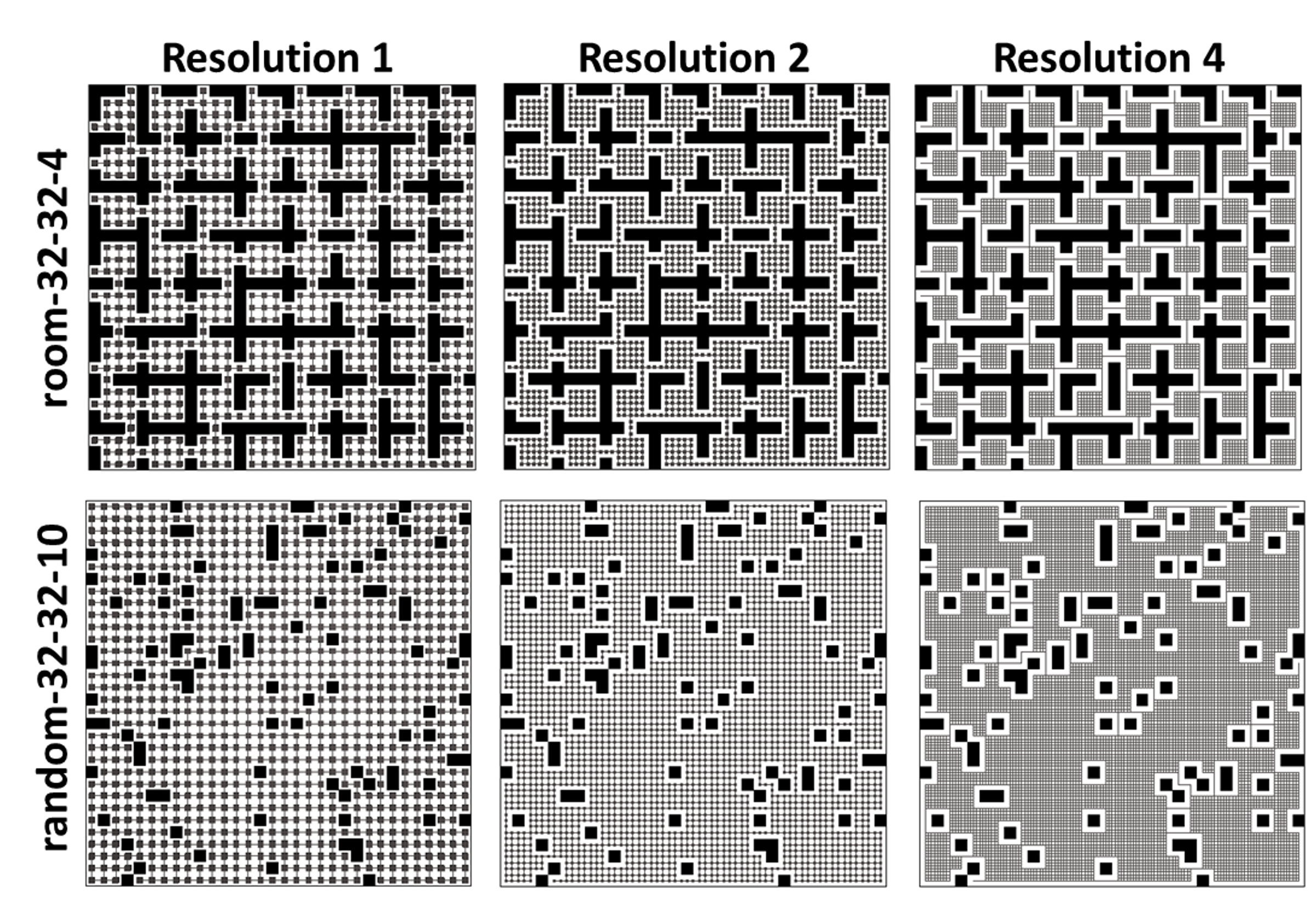}
\caption{The grid roadmap representation graph for two sample environments is shown with varying resolutions of 1, 2, and 4.}
\label{fig:graph_resolution}
\end{figure}

\subsection{Representation Resolution} \label{section:resolution}

The performance of CBS and CBSw/P is significantly influenced by the density and topology of the map. Our grid roadmap-based representation simplifies the inherent complexities of MRMP roadmap representations but also affords the control to enhance the uniform granularity of the representation resolution. The resolutions encompass 1, 2, and 4, where grid roadmap resolution indicates the sampling density within the space while maintaining consistent coverage. Figure \ref{fig:resolution} provides visual examples of these grid resolutions in a sample environment and Figure \ref{fig:graph_resolution} provides visual examples of these grid resolutions in some of our benchmark environments. At the lowest resolution level of 1, vertices are placed at the center of cells in true grid representations, reflecting the planning behavior seen in MAPF grid representations. As the resolution increases, the grid roadmap is subdivided to maintain the same coverage area but with a higher density of vertices. Specifically, a grid roadmap at resolution 1 is divided into 2 and 4 subdivisions for resolution levels 2 and 4, respectively. 

The representation resolution allows us to achieve finer granularity and detail in the representation and facilitate more precise agent movements while maintaining consistent representation coverage. By adjusting the resolution, we can emulate the behavior observed in MRMP roadmaps with higher edge discretization. For instance, when projecting a path at resolution 1 into higher resolutions, as illustrated in Figure \ref{fig:graph_resolution}, the number of states in the path increases correspondingly. This closely resembles the process of increasing edge discretization in an MRMP roadmap representation. In essence, manipulating the representation resolution offers a flexible means to simulate the characteristics of more complex representations, enabling us to study their effects on algorithm performance and behavior. 
 
This versatility in resolution choices allows for a nuanced approach in addressing MRMP challenges. Our grid roadmap-based resolution exploration allows us to gain insight into the challenges faced in MRMP roadmaps without having to consider the nonuniformity of roadmap edge durations. Our hypothesis suggests that CBS is better suited for roadmap representations with less granularity, particularly in scenarios with fewer agents. Conversely, as representations increase with the number of agents and the resolution increases, CBSw/P is anticipated to outperform CBS. This analysis aims to shed light on determining the scalability of MAPF solvers into MRMP problems characterized by larger and more complex representations. By understanding the performance dynamics of CBS and CBSw/P in varying resolutions, we can make informed decisions about their applicability in real-world, intricate environments.

\subsection{Representation Topology}

MAPF and MRMP are sensitive to the topology of an environment, influenced by the robot's shape and size relative to environmental features. Narrow passages restrict movement, while open spaces allow more freedom. While practitioners often develop an intuitive understanding of environmental topology through experience, analyzing the topology of abstract representations remains significantly more challenging.

It is important to distinguish between the topology of the environment and the topology of its representation. \textit{Environment topology} refers to the intrinsic geometric and structural properties of the physical space. \textit{Representation topology}, by contrast, refers to the structure encoded in the chosen abstraction--such as a grid, roadmap, or graph--used to approximate the topology of the planning space (e.g., configuration space). This planning space is determined by both the environment and the robot’s motion capabilities, including its geometry and degrees of freedom. While many representations aim to preserve the topological structure of the planning space so that the representation topology reflects key features of the environment, methods like sampling-based planning can introduce artifacts that distort or incompletely capture this structure, leading to a mismatch between the representation topology and the environment topology.

MAPF algorithms operate on a fixed representation and do not construct it. As such, when applying MAPF to different planning domains, the relevant topology is not that of the environment itself, but of the representation. This distinction is critical: the same environment can yield multiple representations that emphasize different topological features, as illustrated in Figure \ref{fig:mrmp}. Consequently, performance and coordination behavior are shaped more by representation topology than by the physical environment.

In mobile robot settings, the robots typically have low degrees of freedom and simple geometries, allowing structured representations such as grids to align closely with the environment topology, making the relationship between environment topology and representation topology relatively straightforward. In contrast, MRMP for higher-dimensional systems with complex geometries, such as robotic manipulators, typically relies on sampling-based representations, such as Probabilistic Roadmaps (PRM) \cite{l-prmfrpp-98} or Rapidly-exploring Random Trees (RRT) \cite{l-rrtantfpp-98}. These methods sample the free configuration space and connect nearby, collision-free configurations. While dense connections may form in open regions and sparse connections in constrained areas, the resulting roadmaps are still often irregular and introduce artificial topological features that obscure the underlying structure. For high-DoF systems, the configuration space topology may have little to no direct correlation with the physical environment topology, further reinforcing the need to analyze and understand the representation topology itself.

Classical methods for analyzing environment topology, such as Voronoi diagrams and visibility graphs, are less effective for characterizing representation topology. Voronoi diagrams decompose space into regions based on proximity to obstacles, highlighting open spaces and narrow corridors \cite{f-vdadt-17, was-aprmpwsotmaotfs-99, a-vdasoafgds-91}. Visibility graphs connect obstacle vertices with direct lines of sight, capturing similar features \cite{ltw-aafpcfpapo-79, ot-vgalva-01}. Both methods require explicit geometric knowledge of the environment and become impractical in high-dimensional configuration spaces, making them unsuitable for determining representation topology.

\begin{figure}[!t]
\centering
\includegraphics[width=\linewidth]{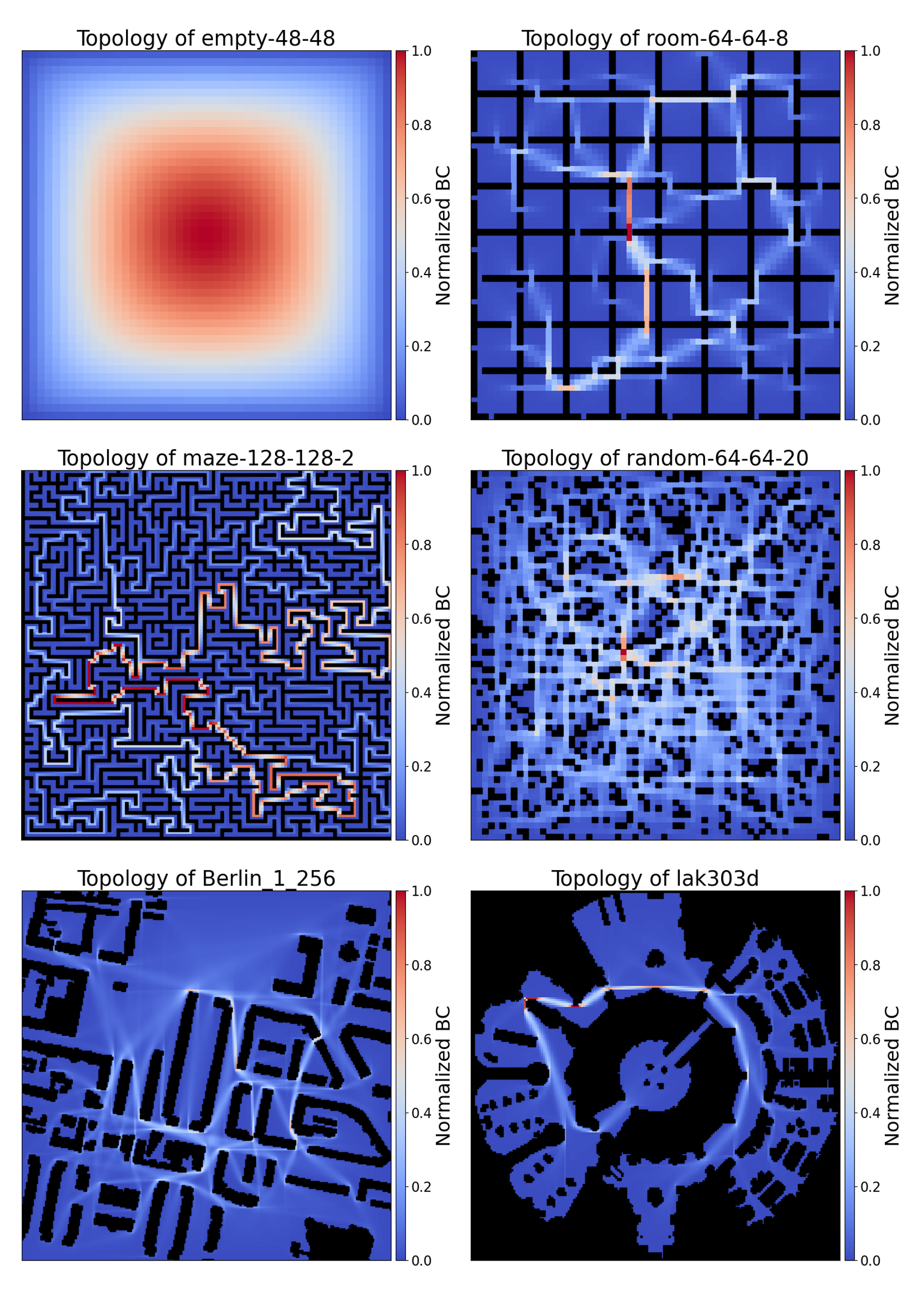}
\caption{Betweenness Centrality visualizations for six environments. Blue indicates low betweenness centrality and red indicates high betweenness centrality. Note that the betweenness centrality is not effective for environments devoid of obstacles.}
\label{fig:top_analysis}
\end{figure}

To address this limitation, betweenness centrality (BC) offers a graph-theoretic approach for inferring topological structure from a given representation \cite{b-afafbc-01}. BC quantifies how frequently a node lies on the shortest paths between all pairs of nodes. Mathematically, the BC of a node $v$ is defined as:
\begin{align*}
    BC(v) = \sum_{s \neq v \neq t} \frac{\sigma(s, t \ | \  v)}{\sigma(s, t)}
\end{align*}
where $\sigma(s, t)$ denotes the number of shortest paths from nodes $s$ and $t$, and $\sigma(s, t\ |\ v)$ is the number of those paths that pass through $v$. This measure reflects a node's importance in maintaining connectivity across the network, making it well-suited for identifying critical topological features such as bottlenecks and narrow passages. High BC values correspond to frequently traversed nodes, while low values indicate regions with multiple alternate routes. 

Figure \ref{fig:top_analysis} illustrates the application of BC across several environments used in our experiments. The complete set of visualizations and the code for replicating this analysis are available in our public GitHub Repository\footref{git_repo}. When applied to grid roadmaps, BC highlights narrow passages as chains of high-centrality nodes and identifies bottlenecks as isolated high-centrality nodes surrounded by low-centrality regions.

\begin{figure}[!t]
\centering
\includegraphics[width=\linewidth]{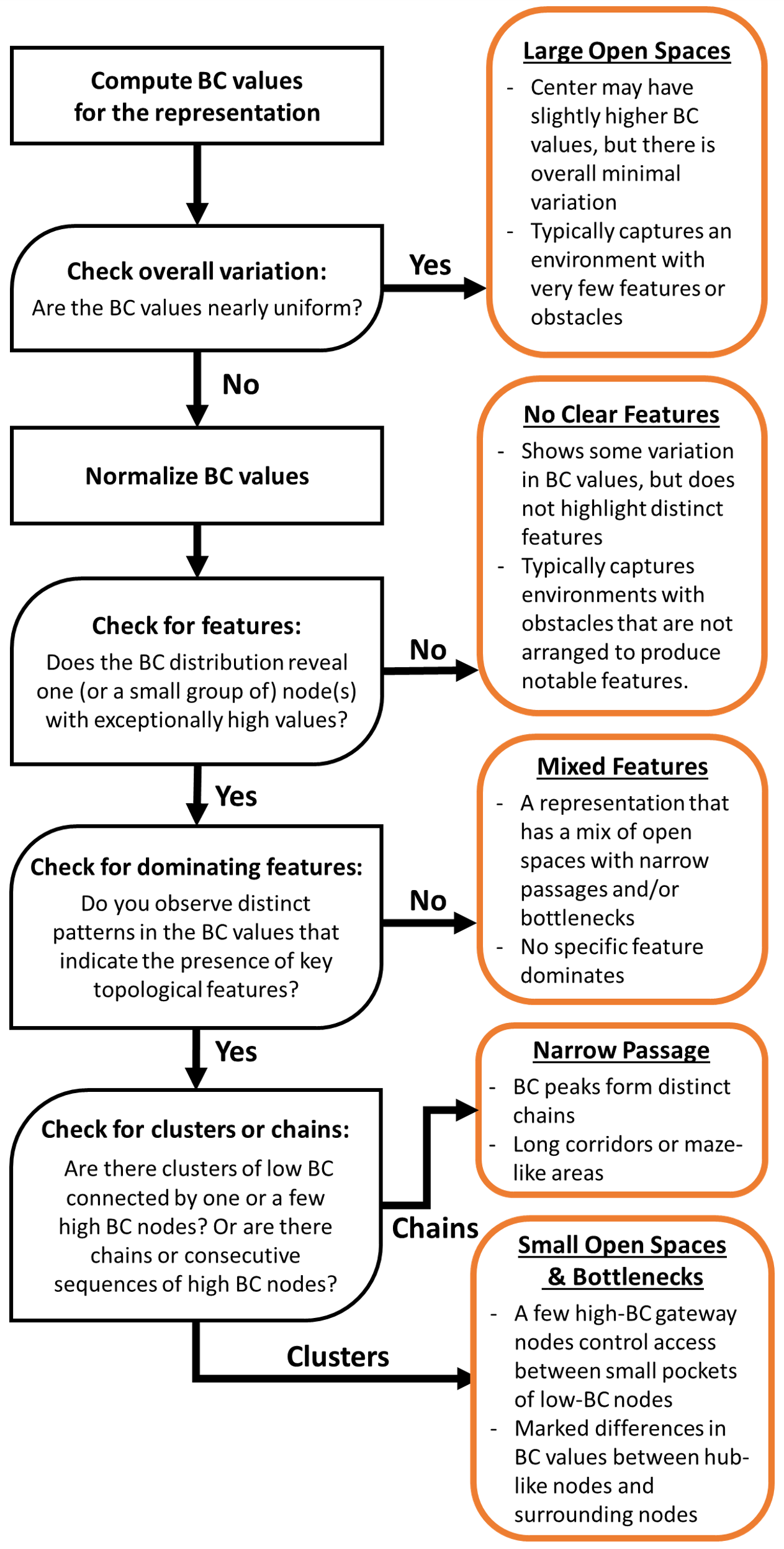}
\caption{A flowchart for identifying the dominant topology within a representation using betweenness centrality.}
\label{fig:topology-flowchart}
\end{figure}

This method performs well across a variety of environments, with the exception of empty ones. In room and maze environments, BC effectively highlights narrow passages and bottlenecks, identifying key locations likely to experience heavy traversal. In randomly scattered environments, BC identifies a few potential narrow passages but does not reveal distinct topological features. Small open spaces are marked by clusters of low-centrality nodes, often connected to other open areas via narrow passages or bottlenecks with high-centrality values.

In small open spaces, there tend to be fewer alternative routes for traversing between different points. This results in higher centrality for the central nodes in these small spaces because more paths must pass through them to connect other points. Thus, in these smaller spaces, a few central nodes may control most of the traffic or flow between others. In large open spaces, there are typically more routes between any two points. As a result, the centrality of individual nodes tends to be lower, even for central locations, because traffic can be distributed over many different paths. Consequently, the BC is spread out across many nodes, with fewer nodes acting as critical intermediaries.

In most environments, BC effectively distinguishes between different topological regions. For instance, in the city map (Berlin\_1\_256) and game map (lak303d) shown in Figure \ref{fig:top_analysis}, BC clearly distinguishes between large open areas and narrow passages. However, in empty environments lacking obstacles, BC tends to highlight the center of the map as the most traversed area, despite the absence of meaningful topological constraints. This limitation arises from the uniformity of the space, which lacks structural features to induce variation in centrality values. Such open environments can be identified by their minimal variance in centrality values prior to normalization. 

BC is a simple and robust tool for identifying topological features within graph-based representations. To support its practical use, we provide a flowchart in Figure \ref{fig:topology-flowchart} to guide users in identifying dominant topological patterns. Based on this method, we categorize the environments used in our experiments into five general types according to their topological features.

\begin{figure}[!t]
\centering
\includegraphics[width=\linewidth]{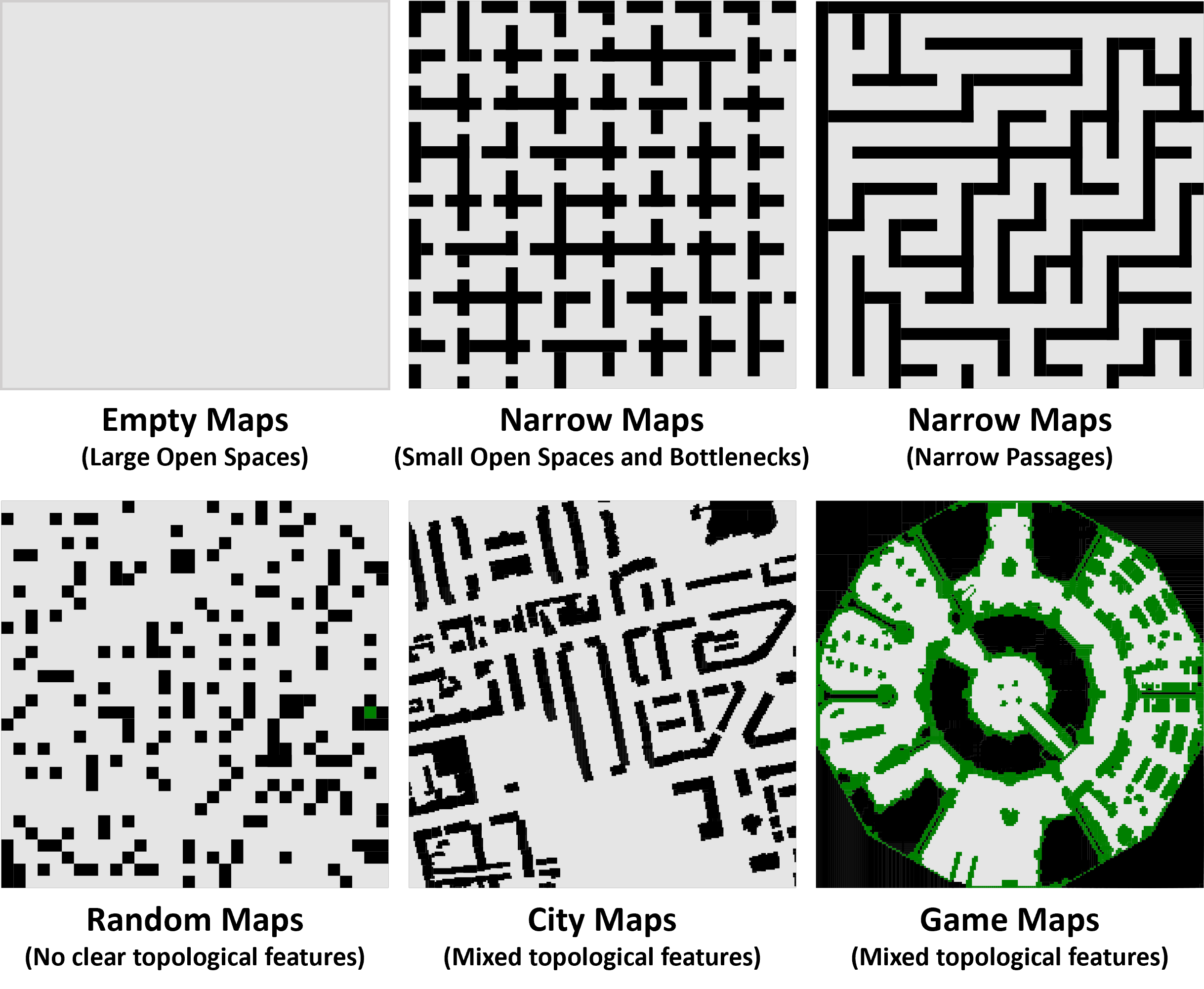}
\caption{All maps fall in one of five groups: empty, random, narrow, cities, and games. These groups are characterized by different topological features. }
\label{fig:maps}
\end{figure}

\subsection{Experimental Setup}

We conduct a comprehensive comparison between CBS and CBSw/P, evaluating their performance in terms of runtime, solution quality (sum-of-costs), and the maximum problem size (number of agents) solved across environments with varying resolutions. Both algorithms are run with a best-first high-level search to ensure that observed differences reflect constraint formulation rather than search strategy. At the low level, we use A* with a 4-neighbor model and sum-of-costs as the objective. Runtime instance plots show the number of CBS and CBSw/P instances solved within the 15-minute limit, plotted against solution time. Additionally, we record the maximum number of agents solved for each map, enabling a direct comparison of scalability. Our experiments compare CBS vs. CBSw/P under these settings and do not claim generality over all conservative versus aggressive methods or over alternative high-level search strategies 

We leverage benchmarks from \cite{ssfkmwlack-mapfdvab-19} and use a total of 27 maps spanning five general categories: empty maps, random maps, narrow maps, real city maps, and video game maps. Figure \ref{fig:maps} provides examples of maps from each category while Table \ref{stats} details the individual maps in each group. These maps are grouped based on topological features that we have identified using betweenness centrality. Empty maps contain no obstacles and are characterized by large open spaces. Random maps contain randomly generated obstacles with no clear topological characterization. Narrow maps contain environments crafted to include narrow passages and bottlenecks; these environments are particularly difficult to solve and include room-like and maze-like maps. Real city maps and video game maps contain a combination of open spaces, narrow passages, and bottlenecks. This benchmark set offers comprehensive coverage of diverse topologies, ranging from artificial environments to real-world MAPF applications through real city and video game maps.

For each map, we execute three different resolutions and 25 scenarios. Each scenario consists of pairs of randomly sampled starts and goals. Following the methodology in \cite{ssfkmwlack-mapfdvab-19}, we begin with a baseline instance of 4 agents. For each subsequent instance, we incrementally add 4 more agents, each with their own start and goal pairs, to the previous instance. This process continues until the solver fails to find a solution for a given instance within the 15-minute runtime limit. We plot the number of solved instances against the time taken. Each algorithm is executed for 25 scenarios, with each scenario tested using resolution 1, resolution 2, and resolution 4 grid roadmaps. This systematic approach allows us to explore the scalability of CBS and CBSw/P across a range of scenarios and grid roadmap resolutions. 

Because we keep the high-level search strategy fixed (best-first), our results highlight the effects motion vs. priority effects rather than strategy effects. We do not include depth-first PBS, bounded-suboptimal CBS, or heuristic/learning-augmented variants that are known to shift runtime and success trade-offs. As a result, the crossover points we report may change under the application of those variants.

\begin{table*}[!t]
% \tiny
\centering
\caption{Graph and Problem Statistics}
{

\resizebox{\textwidth}{!}{%
\begin{tabular}{c|lcrrrrrlclcrrrrr}
 & \multicolumn{1}{c}{\begin{tabular}[c]{@{}c@{}}Map\\ (Size)\end{tabular}} & R & \multicolumn{1}{c}{States} & \multicolumn{2}{c}{\begin{tabular}[c]{@{}c@{}}CBS\\ $-/+$\end{tabular}} & \multicolumn{2}{c}{\begin{tabular}[c]{@{}c@{}}PBS\\ $-/+$\end{tabular}} &  & \multicolumn{1}{c|}{} & \multicolumn{1}{c}{\begin{tabular}[c]{@{}c@{}}Map\\ (Size)\end{tabular}} & R & \multicolumn{1}{c}{States} & \multicolumn{2}{c}{\begin{tabular}[c]{@{}c@{}}CBS\\ $-/+$\end{tabular}} & \multicolumn{2}{c}{\begin{tabular}[c]{@{}c@{}}PBS\\ $-/+$\end{tabular}} \\ \cline{1-8} \cline{10-17} 
\multirow{12}{*}{\centering\arraybackslash   \begin{sideways} Empty \end{sideways}} & \multirow{3}{*}{\begin{tabular}[c]{@{}l@{}}empty-\\ 8-8\\ (8$\times$8)\end{tabular}} & 1 & 64 & 12 & 24 & 12 & 28 &  & \multicolumn{1}{c|}{\multirow{12}{*}{\centering\arraybackslash \begin{sideways} Random   \end{sideways}}} & \multirow{3}{*}{\begin{tabular}[c]{@{}l@{}}random-\\ 32-32-10\\ (32$\times$32)\end{tabular}} & 1 & 922 & 12 & 44 & 32 & 60 \\
 &  & 2 & 225 & 12 & 28 & 24 & 32 &  & \multicolumn{1}{c|}{} &  & 2 & 3185 & 4 & 44 & 28 & 60 \\
 &  & 4 & 841 & 4 & 24 & 20 & 32 &  & \multicolumn{1}{c|}{} &  & 4 & 11575 & 4 & 40 & 28 & 60 \\ \cline{2-8} \cline{11-17} 
 & \multirow{3}{*}{\begin{tabular}[c]{@{}l@{}}empty-\\ 16-16\\ (16$\times$16)\end{tabular}} & 1 & 256 & 12 & 36 & 20 & 40 &  & \multicolumn{1}{c|}{} & \multirow{3}{*}{\begin{tabular}[c]{@{}l@{}}random-\\ 32-32-20\\ (32$\times$32)\end{tabular}} & 1 & 819 & 12 & 40 & 24 & 44 \\
 &  & 2 & 961 & 0 & 40 & 32 & 56 &  & \multicolumn{1}{c|}{} &  & 2 & 2468 & 8 & 40 & 24 & 44 \\
 &  & 4 & 3721 & 0 & 36 & 24 & 56 &  & \multicolumn{1}{c|}{} &  & 4 & 8040 & 4 & 28 & 20 & 40 \\ \cline{2-8} \cline{11-17} 
 & \multirow{3}{*}{\begin{tabular}[c]{@{}l@{}}empty-\\ 32-32\\ (32$\times$32)\end{tabular}} & 1 & 1024 & 12 & 68 & 32 & 80 &  & \multicolumn{1}{c|}{} & \multirow{3}{*}{\begin{tabular}[c]{@{}l@{}}random-\\ 64-64-10\\ (64$\times$64)\end{tabular}} & 1 & 3687 & 20 & 84 & 44 & 88 \\
 &  & 2 & 3969 & 8 & 68 & 32 & 80 &  & \multicolumn{1}{c|}{} &  & 2 & 12830 & 8 & 68 & 28 & 88 \\
 &  & 4 & 15625 & 8 & 40 & 28 & 76 &  & \multicolumn{1}{c|}{} &  & 4 & 46764 & 4 & 56 & 12 & 60 \\ \cline{2-8} \cline{11-17} 
 & \multirow{3}{*}{\begin{tabular}[c]{@{}l@{}}empty-\\ 48-48\\ (48$\times$48)\end{tabular}} & 1 & 2304 & 16 & 84 & 48 & 84 &  & \multicolumn{1}{c|}{} & \multirow{3}{*}{\begin{tabular}[c]{@{}l@{}}random-\\ 64-64-20\\ (64$\times$64)\end{tabular}} & 1 & 3270 & 8 & 52 & 20 & 76 \\
 &  & 2 & 9025 & 16 & 76 & 44 & 100 &  & \multicolumn{1}{c|}{} &  & 2 & 10031 & 8 & 48 & 20 & 48 \\
 &  & 4 & 35721 & 12 & 40 & 16 & 60 &  & \multicolumn{1}{c|}{} &  & 4 & 33225 & 8 & 36 & 20 & 48 \\ \cline{1-8} \cline{10-17} 
\multirow{21}{*}{\centering\arraybackslash   \begin{sideways} Narrow \end{sideways}} & \multirow{3}{*}{\begin{tabular}[c]{@{}l@{}}room-\\ 32-32-4\\ (32$\times$32)\end{tabular}} & 1 & 682 & 8 & 24 & 4 & 28 &  & \multicolumn{1}{c|}{\multirow{27}{*}{\centering\arraybackslash \begin{sideways} Games   \end{sideways}}} & \multirow{3}{*}{\begin{tabular}[c]{@{}l@{}}ht-chantry   \\ (141$\times$162)\end{tabular}} & 1 & 7461 & 12 & 28 & 12 & 28 \\
 &  & 2 & 1902 & 4 & 20 & 4 & 28 &  & \multicolumn{1}{c|}{} &  & 2 & 27912 & 4 & 20 & 4 & 20 \\
 &  & 4 & 5878 & 0 & 20 & 4 & 24 &  & \multicolumn{1}{c|}{} &  & 4 & 107742 & 0 & 16 & 0 & 16 \\ \cline{2-8} \cline{11-17} 
 & \multirow{3}{*}{\begin{tabular}[c]{@{}l@{}}room-\\ 64-64-8 \\ (64$\times$64)\end{tabular}} & 1 & 3232 & 8 & 24 & 8 & 28 &  & \multicolumn{1}{c|}{} & \multirow{3}{*}{\begin{tabular}[c]{@{}l@{}}ht-mans\\ ion-n     \\ (270$\times$133)\end{tabular}} & 1 & 8959 & 4 & 40 & 12 & 40 \\
 &  & 2 & 11090 & 0 & 20 & 8 & 24 &  & \multicolumn{1}{c|}{} &  & 2 & 33376 & 4 & 32 & 0 & 24 \\
 &  & 4 & 40630 & 0 & 16 & 0 & 20 &  & \multicolumn{1}{c|}{} &  & 4 & 128554 & 0 & 20 & 0 & 16 \\ \cline{2-8} \cline{11-17} 
 & \multirow{3}{*}{\begin{tabular}[c]{@{}l@{}}room-\\ 64-64-16 \\ (64$\times$64)\end{tabular}} & 1 & 3646 & 4 & 36 & 8 & 32 &  & \multicolumn{1}{c|}{} & \multirow{3}{*}{\begin{tabular}[c]{@{}l@{}}lak303d\\ (194$\times$194)\end{tabular}} & 1 & 14784 & 4 & 24 & 4 & 24 \\
 &  & 2 & 13585 & 4 & 24 & 8 & 24 &  & \multicolumn{1}{c|}{} &  & 2 & 54938 & 0 & 16 & 0 & 16 \\
 &  & 4 & 52297 & 0 & 16 & 0 & 16 &  & \multicolumn{1}{c|}{} &  & 4 & 211230 & 0 & 8 & 0 & 8 \\ \cline{2-8} \cline{11-17} 
 & \multirow{3}{*}{\begin{tabular}[c]{@{}l@{}}maze-\\ 32-32-2\\ (32$\times$32)\end{tabular}} & 1 & 666 & 4 & 16 & 8 & 20 &  & \multicolumn{1}{c|}{} & \multirow{3}{*}{\begin{tabular}[c]{@{}l@{}}lt-gallows\\ templar     \\ (180$\times$251)\end{tabular}} & 1 & 10021 & 8 & 36 & 12 & 36 \\
 &  & 2 & 1951 & 0 & 16 & 4 & 16 &  & \multicolumn{1}{c|}{} &  & 2 & 37637 & 8 & 36 & 8 & 32 \\
 &  & 4 & 6381 & 0 & 12 & 0 & 12 &  & \multicolumn{1}{c|}{} &  & 4 & 145549 & 0 & 16 & 0 & 20 \\ \cline{2-8} \cline{11-17} 
 & \multirow{3}{*}{\begin{tabular}[c]{@{}l@{}}maze-\\ 32-32-4\\ (32$\times$32)\end{tabular}} & 1 & 790 & 4 & 16 & 4 & 16 &  & \multicolumn{1}{c|}{} & \multirow{3}{*}{\begin{tabular}[c]{@{}l@{}}den312d\\ (81$\times$65)\end{tabular}} & 1 & 2445 & 4 & 28 & 16 & 32 \\
 &  & 2 & 2695 & 0 & 16 & 5 & 16 &  & \multicolumn{1}{c|}{} &  & 2 & 8779 & 4 & 28 & 8 & 28 \\
 &  & 4 & 9853 & 0 & 12 & 4 & 12 &  & \multicolumn{1}{c|}{} &  & 4 & 33105 & 0 & 20 & 4 & 24 \\ \cline{2-8} \cline{11-17} 
 & \multirow{3}{*}{\begin{tabular}[c]{@{}l@{}}maze-\\ 128-128-2\\ (128$\times$128)\end{tabular}} & 1 & 10858 & 0 & 12 & 0 & 12 &  & \multicolumn{1}{c|}{} & \multirow{3}{*}{\begin{tabular}[c]{@{}l@{}}ost003d \\ (194$\times$194)\end{tabular}} & 1 & 13214 & 4 & 32 & 4 & 32 \\
 &  & 2 & 32383 & 0 & 4 & 0 & 4 &  & \multicolumn{1}{c|}{} &  & 2 & 49986 & 0 & 20 & 0 & 20 \\
 &  & 4 & 107437 & 0 & 0 & 0 & 0 &  & \multicolumn{1}{c|}{} &  & 4 & 194168 & 0 & 12 & 0 & 12 \\ \cline{2-8} \cline{11-17} 
 & \multirow{3}{*}{\begin{tabular}[c]{@{}l@{}}maze-\\ 128-128-10\\ (128$\times$128)\end{tabular}} & 1 & 14818 & 0 & 20 & 0 & 20 &  & \multicolumn{1}{c|}{} & \multirow{3}{*}{\begin{tabular}[c]{@{}l@{}}brc202d \\ (481$\times$530)\end{tabular}} & 1 & 43151 & 0 & 16 & 0 & 16 \\
 &  & 2 & 56143 & 0 & 16 & 0 & 16 &  & \multicolumn{1}{c|}{} &  & 2 & 162968 & 0 & 8 & 0 & 8 \\
 &  & 4 & 218317 & 0 & 4 & 0 & 4 &  & \multicolumn{1}{c|}{} &  & 4 & 632432 & 0 & 4 & 0 & 4 \\ \cline{1-8} \cline{11-17} 
\multirow{9}{*}{\centering\arraybackslash   \begin{sideways} Cities \end{sideways}} & \multirow{3}{*}{\begin{tabular}[c]{@{}l@{}}Berlin-\\ 1-256\\ (256$\times$256)\end{tabular}} & 1 & 47536 & 4 & 88 & 24 & 92 &  & \multicolumn{1}{c|}{} & \multirow{3}{*}{\begin{tabular}[c]{@{}l@{}}den520d\\ (257$\times$256)\end{tabular}} & 1 & 28178 & 0 & 48 & 4 & 28 \\
 &  & 2 & 182171 & 4 & 60 & 8 & 48 &  & \multicolumn{1}{c|}{} &  & 2 & 108918 & 0 & 24 & 0 & 24 \\
 &  & 4 & 712615 & 0 & 36 & 0 & 36 &  & \multicolumn{1}{c|}{} &  & 4 & 427970 & 0 & 16 & 0 & 16 \\ \cline{2-8} \cline{11-17} 
 & \multirow{3}{*}{\begin{tabular}[c]{@{}l@{}}Boston-\\ 0-256\\ (256$\times$256)\end{tabular}} & 1 & 47747 & 12 & 64 & 8 & 52 &  & \multicolumn{1}{c|}{} & \multirow{3}{*}{\begin{tabular}[c]{@{}l@{}}w-wound\\ edcoast\\ (578x642)\end{tabular}} & 1 & 34002 & 4 & 20 & 4 & 16 \\
 &  & 2 & 181232 & 4 & 40 & 4 & 32 &  & \multicolumn{1}{c|}{} &  & 2 & 127838 & 0 & 8 & 0 & 8 \\
 &  & 4 & 705218 & 0 & 12 & 0 & 12 &  & \multicolumn{1}{c|}{} &  & 4 & 495024 & 0 & 0 & 0 & 0 \\ \cline{2-8} \cline{10-17} 
 & \multirow{3}{*}{\begin{tabular}[c]{@{}l@{}}Paris-\\ 1-256\\ (256$\times$256)\end{tabular}} & 1 & 47216 & 0 & 68 & 24 & 64 &  &  &  &  &  &  &  &  &  \\
 &  & 2 & 179271 & 0 & 44 & 4 & 44 &  &  &  &  &  &  &  &  &  \\
 &  & 4 & 697685 & 0 & 24 & 0 & 24 &  &  &  &  &  &  &  &  &  \\ \cline{1-8}
\end{tabular}%
}
}
\label{stats}
\end{table*}

\section{Results and Discussion} \label{section:results}

In this section, we discuss how to interpret our plots and synthesize our findings into key takeaways to provide a concise summary of our study. We recommend reading through this section while viewing the flowchart shown in Figure \ref{fig:flowchart}.

\subsection{Plot and Table Analysis}
\begin{figure}[!t]
\centering
\includegraphics[width=.95\linewidth]{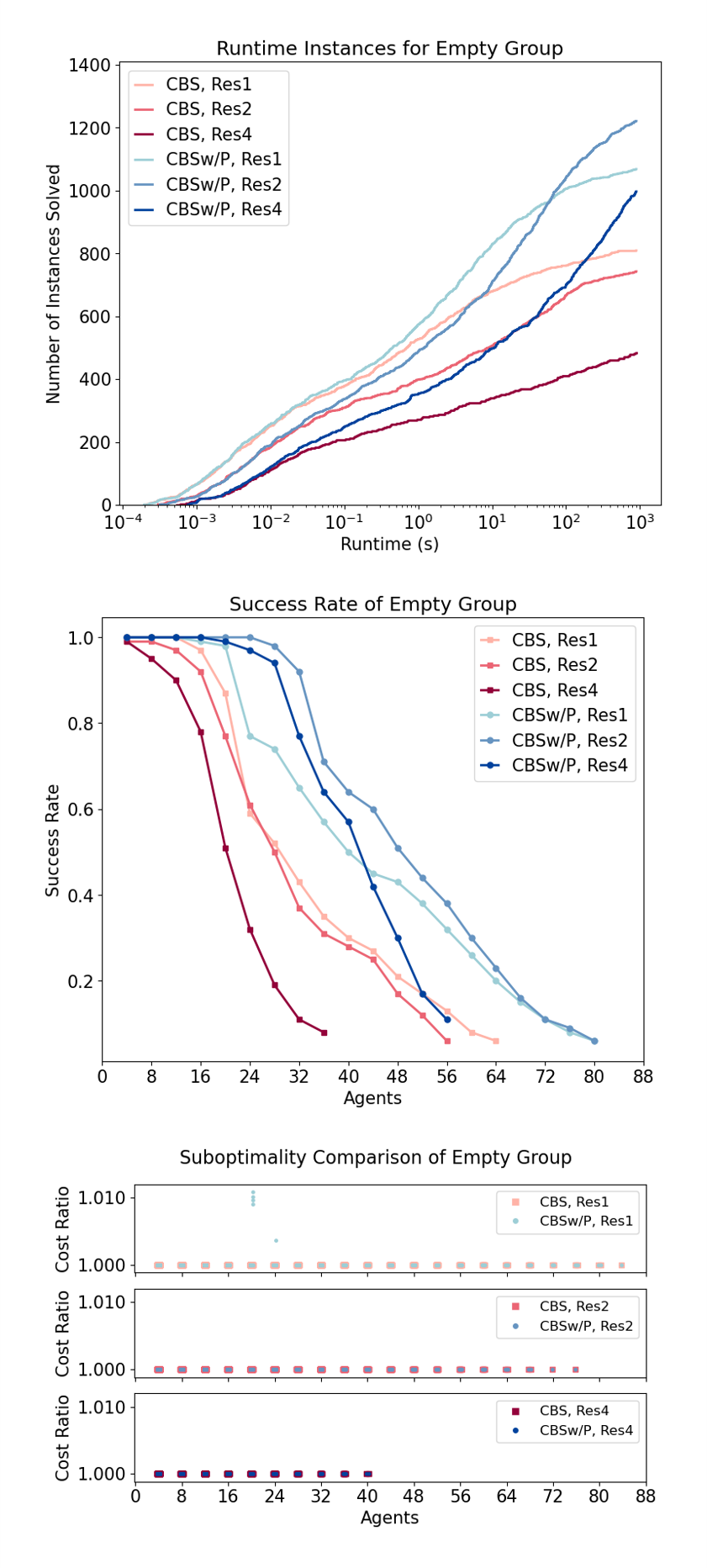}
\caption{Analysis of runtime instances (with varying x-axes), success rate, and cost ratios for the Empty Group. The runtime instance plots display instances ranging from 4 to 80 agents.}
\label{exp:all_empty}
\end{figure}

We compare motion constraints (CBS) and priority constraints (CBSw/P) while holding the high-level search strategy fixed (best-first) so that our findings reflect differences in constraint formulation. Our results are presented in Table \ref{stats} and the plots in  \cref{exp:all_empty,exp:all_narrow,exp:all_random,exp:all_cities,exp:all_games}. Table \ref{stats} provides insights into the size of the environments, the number of states within the state space, the resolution ($R$), and the minimum to maximum problem sizes ($-/+)$ successfully solved by both CBS and CBSw/P. The main takeaway from this table should be the increase in states as resolution increases and the maps in each grouping. The plots in  \cref{exp:all_empty,exp:all_narrow,exp:all_random,exp:all_cities,exp:all_games} offer an average perspective based on map groupings and present runtime instances, average success rates, and sum-of-costs ratios, facilitating a comparison between CBSw/P and CBS. For a detailed exploration of the same statistics on an individual map basis, along with raw data, please refer to our public GitHub repository\footref{git_repo}.

These plots are averaged over map groupings and scenarios. Scenarios exhibit varying difficulty due to random generation, which allows us to draw conclusions independent of problem difficulty. Some scenarios are easier due to shorter or non-conflicting paths, while others pose increased difficulty with longer and conflicting paths. Map groupings are based on topological features, with each grouping containing maps of different sizes. By grouping maps according to topology, we can make conclusions that are independent of environment size. These averaged grouped plots effectively highlight key trends relevant to our analysis on the influence of environment representation topology and representation resolution, while minimizing the impact of environment size and problem difficulty variations.

For our runtime analysis, we present runtime instance plots, which show the number of instances solved within the 15-minute limit across all maps and scenarios in a given group. In these experiments, the number of robots was increased in increments of four agents, so each successive solved instance represents a configuration with four more robots than the previous one. These plots avoid the skew introduced by averaging runtimes, where varying scenario difficulty can produce irregular peaks and dips as problem sizes increase and success rates decline. By counting solved instances directly, runtime instance plots provide a clearer view of algorithm performance and scalability. The x-axis of these plots is logarithmic.

We calculate the success rate by dividing the total number of solved scenarios by the total number of possible scenarios across all maps in a given group. Lastly, for our cost ratio plots, we compute the cost ratio between CBSw/P and CBS for scenarios where both algorithms successfully solved the problem. If either algorithm fails to solve a scenario, the cost ratio is not reported, as it cannot be calculated.

\subsection{Analysis of Results} \label{section:analysis_results}

In this subsection, we consolidate our findings by examining the plots corresponding to our map groupings and elucidating the decision-making process outlined in our flowchart (Figure  \ref{fig:flowchart}). Our discussion summarizes vanilla CBS and CBSw/P under our evaluation settings and should be viewed as a first-pass heuristic rather than a universal prescription for all conservative and aggressive methods.

\subsubsection{Large Open Space Environments}

Figure \ref{exp:all_empty} displays the runtime instance, average success rate, and cost ratios for the empty group of maps, characterized by representations containing vast open spaces devoid of interesting features. This group poses relatively easier challenges compared to other map groupings due to the absence of obstructive elements. 

Our analysis consistently shows that CBSw/P achieves higher success rates and comparable or quicker runtimes than CBS across different problem sizes and resolutions. This superior performance comes with a slight compromise in solution quality, as indicated by a maximum suboptimal cost of only about 1.015 times the optimal. In representations with large open spaces, CBSw/P matches or exceeds CBS in both success rate and runtime, irrespective of the problem size or resolution. Therefore, in such settings, we suggest utilizing aggressive constraints for both MAPF and MRMP tasks. 

The aggressive nature of CBSw/P's constraints aids in the efficient discovery of team solutions, particularly with increasing numbers of agents. Moreover, as resolution increases, the benefits of using aggressive constraints become more pronounced. Thus, regardless of the problem's scale or representation resolution, we recommend using aggressive constraints because they offer superior scalability without any significant degradation in solution quality. Notably, CBSw/P achieves better scalability with a resolution 2 grid roadmap compared to a resolution 1 grid roadmap. By increasing the number of states in the representation, agents can make smaller movements, which is particularly beneficial in scenarios with higher agent densities. While the resolution 2 grid roadmap results in higher runtimes due to the increased number of possible states, it also offers improved scalability with respect to problem size.

The advantage of aggressive constraints in open spaces lies in their emphasis on rapid exploration of the search space. They navigate these areas efficiently without being impeded by numerous conflicts, unlike conservative constraints, which may spend excessive time analyzing unnecessary conflicts. Aggressive constraints perform particularly well in scenarios where high levels of coordination are not required. As a result, their inherent characteristics align naturally with open-space representations, making them a more suitable and efficient choice for solving problems in such contexts compared to conservative constraints.

\begin{figure}[!t]
\centering
\includegraphics[width=.95\linewidth]{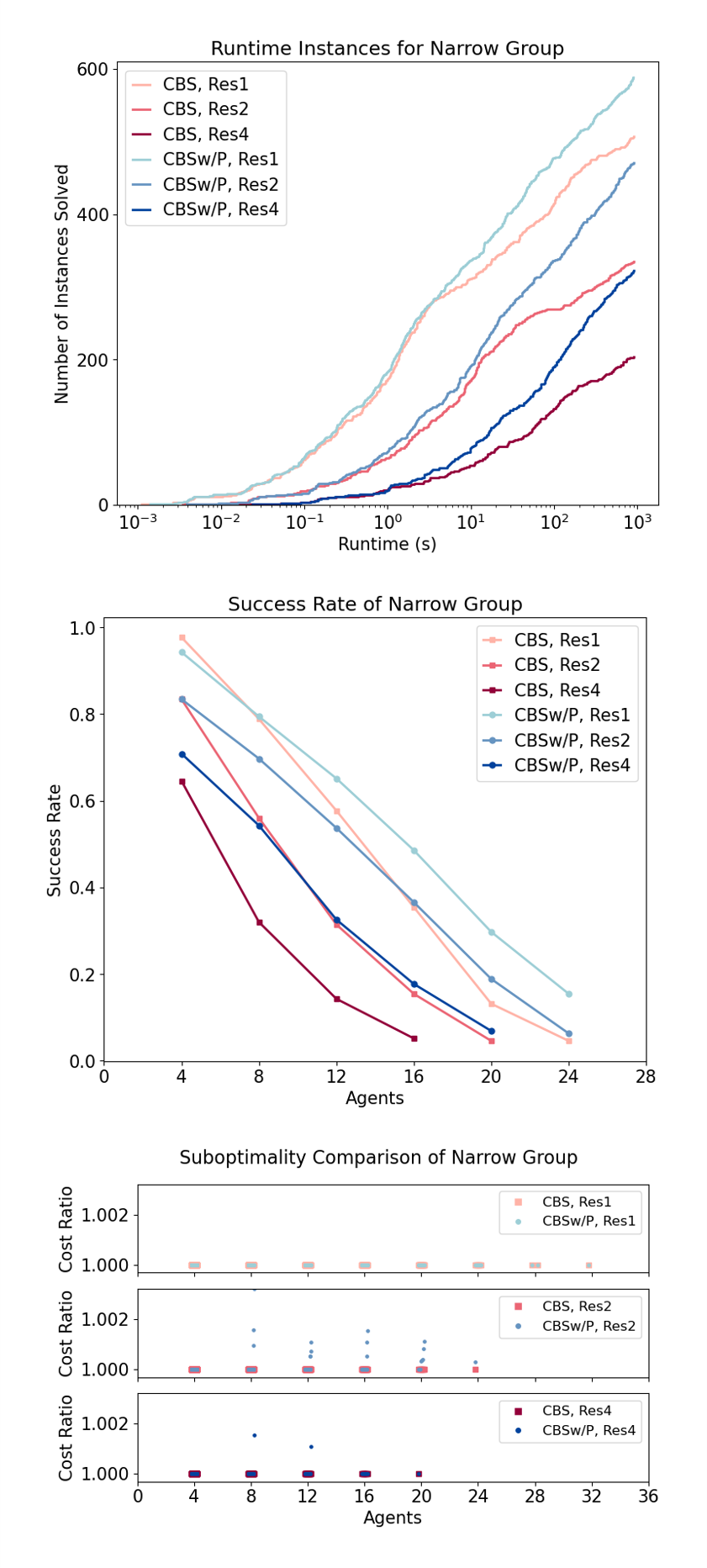}
\caption{Analysis of runtime instances (with varying x-axes), success rate, and cost ratios for the Narrow Group. The runtime instance plots display instances ranging from 4 to 24 agents. }
\label{exp:all_narrow}
\end{figure}

\subsubsection{Narrow Environments}

Figure \ref{exp:all_narrow} presents the results for the narrow group of maps, which are characterized by numerous narrow passages or bottlenecks coupled with small open spaces. These environments have historically presented formidable challenges due to obstacle configurations that obstruct the traversal of multiple robots, significantly complicating coordination efforts. 

In navigating representations with narrow passages, our analysis begins by considering problem size and representation resolution. For scenarios involving a small number of agents, conservative constraints prove advantageous in ensuring more coordinated behavior between agents. Particularly in cases where both the number of agents is small and the representation resolution is low, conservative constraints demonstrate comparable runtimes and success rates to aggressive constraints. 

However, as the number of agents navigating congested regions increases, conservative constraints struggle to find solutions efficiently due to their meticulous constraint application, leading to longer runtimes. In contrast, the scalability of aggressive constraints becomes increasingly evident as problem size grows, making them a more practical choice for larger-scale scenarios, regardless of representation resolution, runtime considerations, or completeness trade-offs.

\begin{figure}[!t]
\centering
\includegraphics[width=.95\linewidth]{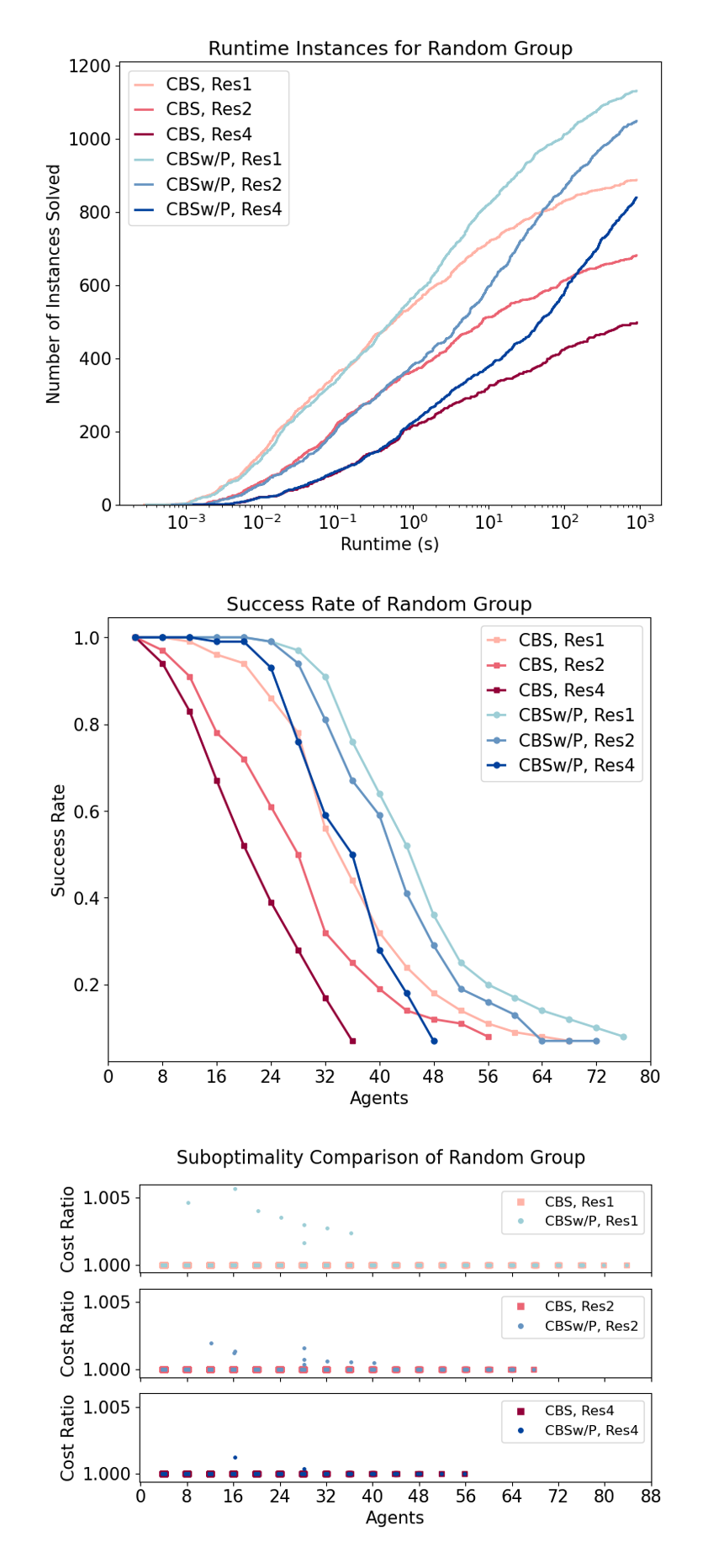}
\caption{Analysis of runtime instances (with varying x-axes), success rate, and cost ratios for the Random Group. The runtime instance plots display instances ranging from 4 to 76 agents.}
\label{exp:all_random}
\end{figure}

When evaluating representation resolution alongside a relatively small number of agents, the choice between constraint types depends on its impact 
on planning time. In MRMP instances with high-resolution representations, aggressive constraints can increase the likelihood of finding a solution by reducing runtime, which may help avoid resource exhaustion. However, aggressive constraints are often incomplete and may not always guarantee a valid solution. In contrast, with lower representation resolutions, conservative constraints provide the advantage of ensuring paths with more coordinated behavior while maintaining runtimes comparable to aggressive constraints. However, we recommend choosing between the two based on the desired completeness and planning time, depending on the specific needs of your problem.

This analysis underscores the intricate relationship between problem attributes, solution strategies, and constraint selection. While conservative constraints excel in ensuring precision, completeness, and optimality, particularly in smaller-scale scenarios, the scalability and exploratory nature of aggressive constraints make them better suited for complex problem instances characterized by numerous agents. Aligning constraint selection with problem attributes enables practitioners to navigate multi-agent systems effectively, optimizing efficiency across 
diverse representations.

\subsubsection{Random Environments}

Random environments are characterized by the arbitrary distribution of obstacles, lacking any predominant topological feature. Figure \ref{exp:all_random} showcases the results derived from our analysis of these random environments. The absence of specific topological features in these representations poses challenges in predicting optimal strategies.

Aggressive constraints offer a practical approach for navigating random representations by swiftly eliminating similar paths, thereby improving computational efficiency. This mirrors their effectiveness in open-space representations, where scalability is often a priority. Since representations capturing random environments typically require low levels of coordination, aggressive constraints excel in these scenarios. However, this comes at the cost of completeness. When guaranteeing a valid solution is a priority, conservative constraints provide an alternative. However, applying conservative constraints in representations with inherently low coordination requirements generally increases computational demands, limiting scalability. As a result, aggressive constraints remain the preferable choice in most cases, as they significantly enhance the likelihood of finding viable solutions within a reasonable timeframe.

In the broader context of MAPF and MRMP, aggressive constraints are generally the optimal choice for representations lacking distinguishing features, as they strike a balance between computational efficiency and solution feasibility. For users prioritizing completeness, particularly in smaller problem instances or when working with low-resolution roadmaps, conservative constraints can be a viable option. Ultimately, the choice of constraints should align with the user’s specific performance goals and the demands of the task at hand.

\subsubsection{Mixed Environments}

\begin{figure}[!t]
\centering
\includegraphics[width=.95\linewidth]{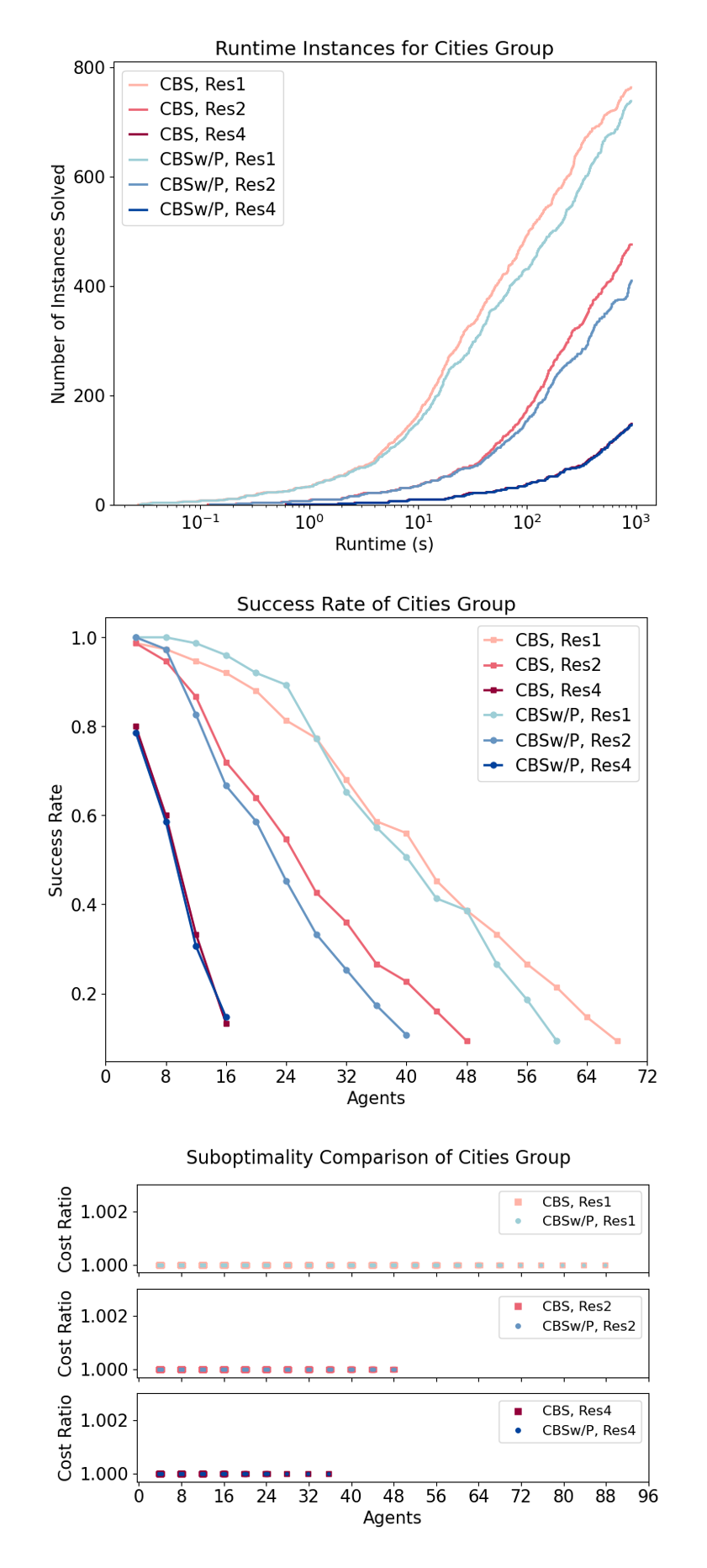}
\caption{Analysis of runtime instances (with varying x-axes), success rate, and cost ratios for the Cities Group. The runtime instance plots display instances ranging from 4 to 68 agents}
\label{exp:all_cities}
\end{figure}

\begin{figure}[!t]
\centering
\includegraphics[width=.95\linewidth]{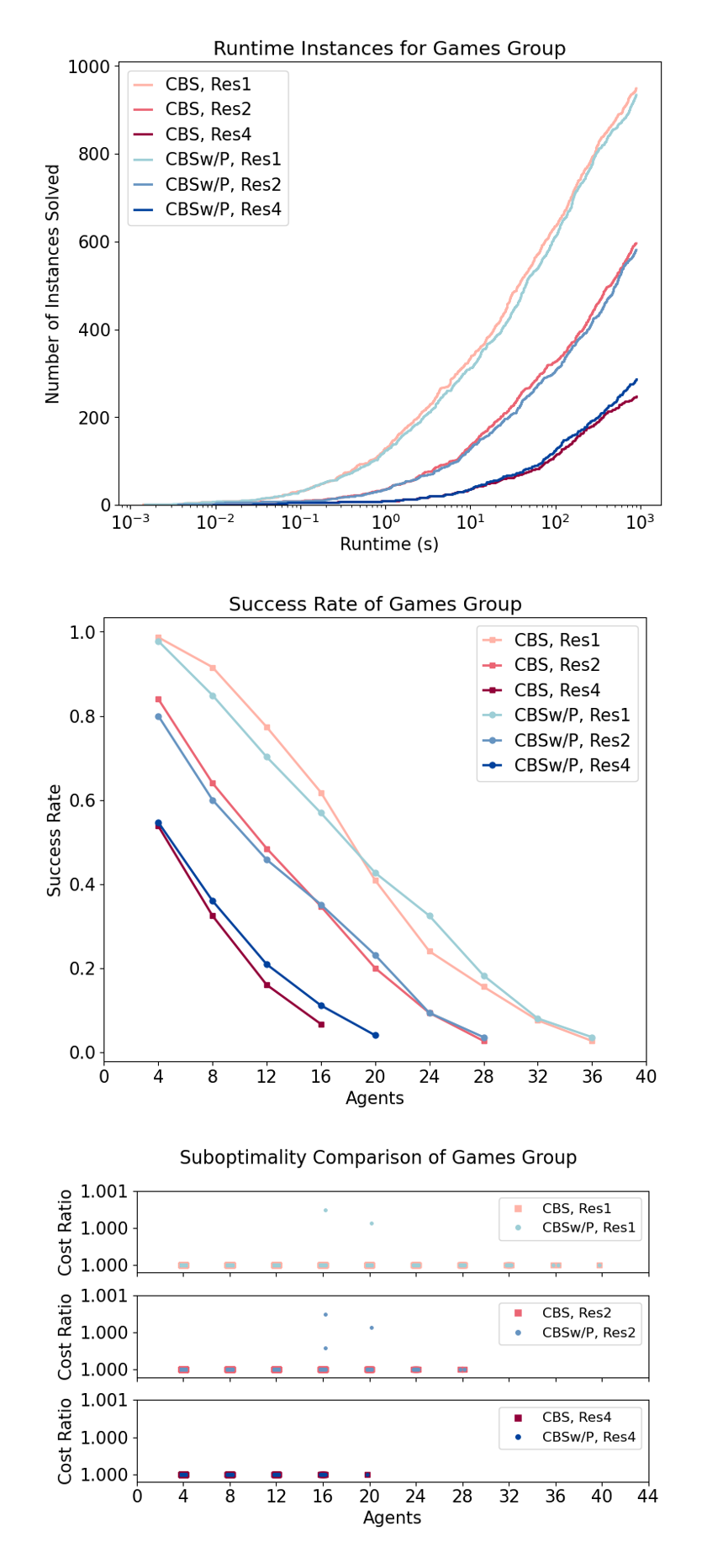}
\caption{Analysis of runtime instances (with varying x-axes), success rate, and cost ratios for the Games Group. The runtime instance plots display instances ranging from 4 to 36 agents.}
\label{exp:all_games}
\end{figure}

Mixed environments encompass various topological features such as open spaces, narrow passages, and bottlenecks. The cities and games groupings exemplify representations that amalgamate these diverse characteristics and their results are shown in Figures \ref{exp:all_cities} and \ref{exp:all_games}, respectively. 

In terms of runtime and success rates, conservative constraints demonstrate comparable or superior performance to aggressive constraints as problem size and representation resolution increase. Conservative constraints provide slightly higher success rates with lower planning times, regardless of problem size or resolution. Therefore, in representations with diverse topological features, we recommend using conservative constraints over aggressive ones.

Conservative constraints perform well in such representations by efficiently identifying and resolving conflicts with a few targeted constraints. Their ability to enforce more coordinated behavior with a few targeted constraints allows conflicts to be addressed quickly, making them particularly advantageous in representations where different regions require varying levels of precision or where conflict paths demand more nuanced conflict resolution. Given their consistent performance, often matching or exceeding that of aggressive constraints, conservative constraints are the preferable choice for mixed topological representations. 

While aggressive constraints can achieve similar results, they tend to incur higher runtimes. As the number of agents increases, the placement of numerous priority constraints to resolve localized conflicts can lead to excessive iterations of planning and validation. Although priority constraints remain a viable approach, the completeness guarantees of conservative constraints make them a more reliable and effective option in comparison. 

\subsection{Key Takeaways} \label{section:takeaways}

Our study highlights the critical role of coordination requirements in selecting appropriate constraints for constraint-based search. When representations impose high coordination demands, conservative constraints are often preferable, as they methodically resolve conflicts and preserve completeness. Conversely, in scenarios with lower coordination requirements, aggressive constraints provide a more efficient alternative by rapidly eliminating conflicts and reducing planning overhead.

Our comparisons fix the high-level search to best-first in order to isolate the effects of constraint formulation; other strategies (e.g., depth-first PBS, bounded suboptimal variants, etc.) can shift success rates and scalability. The heuristics and guidelines we provide are intended as a first-pass reference for practitioners when selecting among constraint-based methods. Another thing to note is that our betweenness-centrality topology measure is informative on many maps but less reliable in empty environments, so trends on empty maps should be interpreted with caution. Finally, while the grid roadmap discretization reduces representation artifacts, it is still an abstraction of  continuous MRMP spaces. With these considerations in mind, we now summarize our findings from comparing vanilla CBS with CBSw/P. 

In mixed representations containing localized regions of dense agent interaction, conservative constraints excel by focusing conflict resolution precisely where it is needed. In contrast, open environments or those lacking distinct topological features benefit from aggressive constraints, which expedite solution discovery by minimizing coordination requirements and pruning large portions of the search space.

However, in representations featuring narrow passages or bottlenecks, further considerations arise. Although conservative constraints offer more coordinated behavior, they can lead to exponential growth in the constraint tree. This occurs because conservative constraints may repeatedly reintroduce and resolve conflicts between the same agent pairs across different branches during replanning iterations. In contrast, aggressive constraints avoid this redundancy by ensuring that each agent pair conflicts at most once within any given branch. This property allows aggressive constraints to limit the branching factor and reduce search complexity, especially in highly coupled environments or as the number of agents increases. Nevertheless, due to their incompleteness, aggressive constraints may struggle in topologically constrained settings, where more conservative conflict resolution is necessary to guarantee solution validity.

The insights from our previous subsection are summarized into an easily digestible decision flowchart, shown in Figure \ref{fig:flowchart}. In this subsection, we will generally follow this flowchart to have a high-level discussion of the results and summarize key takeaways. 

\subsubsection{Representation Topology}

Our observations indicate that representations with large open spaces or a lack of distinct features benefit from aggressive constraints due to their scalability and runtime advantages. In contrast, for representations with a mix of features, conservative constraints are recommended, as they guarantee high-quality solutions while offering comparable or superior scalability and runtime performance. For environment representations dominated by narrow passages or small open spaces with bottlenecks, both the problem size and representation resolution should guide the choice between aggressive and conservative constraints.

\subsubsection{Scale and Granularity Factors}

For settings with large problem sizes or high representation resolutions, aggressive constraints are preferable to conservative constraints due to their superior scalability and computational efficiency. By imposing fewer constraints, they can identify valid paths more quickly, albeit at the cost of completeness. Nevertheless, they remain effective for solving large-scale problems in dense representations with limited computational resources. In contrast, employing conservative constraints in such contexts is less advisable, as the need to specify exact states can result in an excessive number of constraints, increasing computational overhead and delaying solution discovery. For scenarios with smaller problem sizes and lower representation resolutions, the choice of constraints should be guided by the planner's desired balance between runtime efficiency and solution completeness. 

Problem size can be determined by agent density, which refers to the concentration of agents within the traversable free space of an environment. This measure quantifies how crowded an environment is with agents. As a general guideline, ``many agents" typically refers to problems with an agent density greater than 10–20\% of the traversable free space. However, since constraint-based search algorithms scale based on the number of conflicts encountered, a rough estimate of whether a problem has many or few agents is sufficient for the purposes of this flowchart.

\subsubsection{Execution Performance Metrics}

Lastly, the choice of constraints should be guided by the planner's desired runtime performance and completeness requirements. Notably, solution quality should not be a primary factor in this decision, as the differences in solution optimality between constraint types are typically negligible. Instead, the focus should be on balancing computational efficiency and the ability to guarantee a solution. If computational resources are limited or completeness is not a strict requirement, aggressive constraints are preferable due to their faster planning times and scalability. Conversely, if guaranteeing a valid solution is critical and longer planning times are acceptable, conservative constraints should be used to ensure completeness at the cost of increased computational overhead.

\section{Conclusion}

This paper investigates how constraint formulation influences the behavior of constraint-based search for MAPF and MRMP. We compared vanilla CBS and CBSw/P under a fixed best-first high-level search. Our results show that environments with high coordination demands tend to favor conservative constraints, which systematically resolve conflicts and preserve completeness, whereas environments with lower coordination demands often benefit from aggressive constraints, which reduce repeated conflict expansion and planning overhead. 

We synthesize these findings into the decision flowchart in Figure \ref{fig:flowchart}, offering a practical first-pass guide for method selection in MAPF and MRMP applications. These conclusions are limited to the methods and settings studied and should not be interpreted as universal prescriptions for all conservative versus aggressive methods. Alternative high-level strategies and modern enhancements may shift success rates and scalability. Future work could build on this preliminary study by factorially separating constraint formulation from search strategy, incorporating state-of-the-art variants, and broadening topology characterization beyond betweenness centrality and grid-roadmap abstractions. 

Looking ahead, advancements in constraint-based search should account for representation features and application-specific requirements. Designing universally flexible MAPF algorithms remains an appealing but difficult challenge. Effective algorithm development must weigh solver design against target environments, problem scales, resolution needs, and the trade-offs between solution quality and efficiency. Although this study did not address noise or uncertainty, we expect the presented insights to extend to such scenarios, which warrant further investigation. By framing constraints as conservative or aggressive, developers can make more informed design choices, ultimately broadening the effectiveness and applicability of MAPF algorithms.

\section*{Acknowledgment}\label{sec:ack}
This work was supported in part by the U.S. National Science Foundation's ``Expeditions: Mind in Vitro: Computing with Living Neurons"  under award No. IIS-2123781, and by the IBM-Illinois Discovery Accelerator Institute and the Center for Networked Intelligent Components and Environments (C-NICE) at the University of Illinois. 

The work of Hannah Lee was supported in part by the National Science Foundation Graduate Research Fellowship under Grant No. 2020297899. Any opinion, findings, and conclusions or recommendations expressed in this material are those of the authors and do not necessarily reflect the views of the National Science Foundation. 

Marco Morales is supported in part by the  Asociaci\'on Mexicana de Cultura A.C. 
\bibliographystyle{ieeetr}
\bibliography{./references.bib}

@string{ICRA="Proc.\ {IEEE} Int.\ Conf.\ Robot.\ Autom.\ ({ICRA})"}

@string{IROS="Proc.\ {IEEE} Int.\ Conf.\ Intel.\ Rob.\ Syst.\ ({IROS})"}

@inproceedings{ssfkmwlack-mapfdvab-19,
  title={Multi-agent pathfinding: Definitions, variants, and benchmarks},
  author={Stern, Roni and Sturtevant, Nathan and Felner, Ariel and Koenig, Sven and Ma, Hang and Walker, Thayne and Li, Jiaoyang and Atzmon, Dor and Cohen, Liron and Kumar, TK and others},
  booktitle={Proceedings of the International Symposium on Combinatorial Search},
  volume={10},
  number={1},
  pages={151--158},
  year={2019}
}

@inproceedings{ssfsb-aecothomapfutmatsoco-16,
  title={An empirical comparison of the hardness of multi-agent path finding under the makespan and the sum of costs objectives},
  author={Surynek, Pavel and Felner, Ariel and Stern, Roni and Boyarski, Eli},
  booktitle={Proceedings of the International Symposium on Combinatorial Search},
  volume={7},
  number={1},
  pages={145--146},
  year={2016}
}

@inproceedings{fssbgsws-sbosftmapfpsac-17,
  title={Search-based optimal solvers for the multi-agent pathfinding problem: Summary and challenges},
  author={Felner, Ariel and Stern, Roni and Shimony, Solomon and Boyarski, Eli and Goldenberg, Meir and Sharon, Guni and Sturtevant, Nathan and Wagner, Glenn and Surynek, Pavel},
  booktitle={Proceedings of the International Symposium on Combinatorial Search},
  volume={8},
  number={1},
  pages={29--37},
  year={2017}
}

@inproceedings{hlw-agffaptmsa-98,
  title={A general framework for assembly planning: The motion space approach},
  author={Halperin, Dan and Latombe, Jean-Claude and Wilson, Randall H},
  booktitle={Proceedings of the fourteenth annual symposium on Computational geometry},
  pages={9--18},
  year={1998}
}

@book{nb-toaraap-93,
  title={Theory of automatic robot assembly and programming},
  author={Nnaji, Bartholomew O},
  year={1993},
  publisher={Springer Science \& Business Media}
}

@inproceedings{ra-bbep-10,
  title={Behavior-based evacuation planning},
  author={Rodriguez, Samuel and Amato, Nancy M},
  booktitle={2010 IEEE International Conference on Robotics and Automation},
  pages={350--355},
  year={2010},
  organization={IEEE}
}

@article{ba-bbfcfmt-98,
  title={Behavior-based formation control for multirobot teams},
  author={Balch, Tucker and Arkin, Ronald C},
  journal={IEEE transactions on robotics and automation},
  volume={14},
  number={6},
  pages={926--939},
  year={1998},
  publisher={IEEE}
}

@article{tpk-ltfs-04,
  title={Leader-to-formation stability},
  author={Tanner, Herbert G and Pappas, George J and Kumar, Vijay},
  journal={IEEE Transactions on robotics and automation},
  volume={20},
  number={3},
  pages={443--455},
  year={2004},
  publisher={IEEE}
}

@article{kh-ppfpimf-06,
  title={Path planning for permutation-invariant multirobot formations},
  author={Kloder, Stephen and Hutchinson, Seth},
  journal={IEEE Transactions on Robotics},
  volume={22},
  number={4},
  pages={650--665},
  year={2006},
  publisher={IEEE}
}

@inproceedings{lsmfkk-maifice-20,
  title={Moving agents in formation in congested environments},
  author={Li, Jiaoyang and Sun, Kexuan and Ma, Hang and Felner, Ariel and Kumar, TK and Koenig, Sven},
  booktitle={Proceedings of the International Symposium on Combinatorial Search},
  volume={11},
  number={1},
  pages={131--132},
  year={2020}
}

@inproceedings{lwcycl-mfifadhlatmamt-21,
  title={Moving Forward in Formation: A Decentralized Hierarchical Learning Approach to Multi-Agent Moving Together},
  author={Liu, Shanqi and Wen, Licheng and Cui, Jinhao and Yang, Xuemeng and Cao, Junjie and Liu, Yong},
  booktitle={2021 IEEE/RSJ International Conference on Intelligent Robots and Systems (IROS)},
  pages={4777--4784},
  year={2021},
  organization={IEEE}
}

@inproceedings{bpssk-ostaapffmarap-20,
  title={Optimal sequential task assignment and path finding for multi-agent robotic assembly planning},
  author={Brown, Kyle and Peltzer, Oriana and Sehr, Martin A and Schwager, Mac and Kochenderfer, Mykel J},
  booktitle={2020 IEEE International Conference on Robotics and Automation (ICRA)},
  pages={441--447},
  year={2020},
  organization={IEEE}
}

@article{fbkt-apptcmrl-00,
  title={A probabilistic approach to collaborative multi-robot localization},
  author={Fox, Dieter and Burgard, Wolfram and Kruppa, Hannes and Thrun, Sebastian},
  journal={Autonomous robots},
  volume={8},
  pages={325--344},
  year={2000},
  publisher={Springer}
}

@inproceedings{rdjj-mfwtofar-95,
  title={Moving furniture with teams of autonomous robots},
  author={Rus, Daniela and Donald, Bruce and Jennings, Jim},
  booktitle={Proceedings 1995 IEEE/RSJ International Conference on Intelligent Robots and Systems. Human Robot Interaction and Cooperative Robots},
  volume={1},
  pages={235--242},
  year={1995},
  organization={IEEE}
}

@inproceedings{yl-saioomrppog-13,
  title={Structure and intractability of optimal multi-robot path planning on graphs},
  author={Yu, Jingjin and LaValle, Steven},
  booktitle={Proceedings of the AAAI Conference on Artificial Intelligence},
  volume={27},
  number={1},
  pages={1443--1449},
  year={2013}
}

@inproceedings{pbs,
  title={Searching with consistent prioritization for multi-agent path finding},
  author={Ma, Hang and Harabor, Daniel and Stuckey, Peter J and Li, Jiaoyang and Koenig, Sven},
  booktitle={Proceedings of the AAAI conference on artificial intelligence},
  volume={33},
  number={01},
  pages={7643--7650},
  year={2019}
}

@article{cbs,
  title={Conflict-based search for optimal multi-agent pathfinding},
  author={Sharon, Guni and Stern, Roni and Felner, Ariel and Sturtevant, Nathan R},
  journal={Artificial Intelligence},
  volume={219},
  pages={40--66},
  year={2015},
  publisher={Elsevier}
}

@article{s-mapfao-19,
  title={Multi-agent path finding--an overview},
  author={Stern, Roni},
  journal={Artificial Intelligence: 5th RAAI Summer School, Dolgoprudny, Russia, July 4--7, 2019, Tutorial Lectures},
  pages={96--115},
  year={2019},
  publisher={Springer}
}

@techreport{lsj-asotmapp-21,
  title={A survey of the multi-agent pathfinding problem},
  author={Lejeune, Erwin and Sarkar, Sampreet and Jezequel, Lo{\i}g},
  year={2021},
  institution={Technical Report, Accessed July}
}

@inproceedings{bs-osbafmapfwtsoco-19,
  title={On SAT-based approaches for multi-agent path finding with the sum-of-costs objective},
  author={Bart{\'a}k, Roman and {\v{S}}vancara, Ji{\v{r}}{\'\i}},
  booktitle={Proceedings of the International Symposium on Combinatorial Search},
  volume={10},
  number={1},
  pages={10--17},
  year={2019}
}

@inproceedings{sfsb-esatmapfutsoco-16,
  title={Efficient SAT approach to multi-agent path finding under the sum of costs objective},
  author={Surynek, Pavel and Felner, Ariel and Stern, Roni and Boyarski, Eli},
  booktitle={Proceedings of the twenty-second european conference on artificial intelligence},
  pages={810--818},
  year={2016}
}

@inproceedings{cs-daiomapfasst-21,
  title={DPLL (MAPF): an Integration of Multi-Agent Path Finding and SAT Solving Technologies},
  author={{\v{C}}apek, Martin and Surynek, Pavel},
  booktitle={Proceedings of the International Symposium on Combinatorial Search},
  volume={12},
  number={1},
  pages={153--155},
  year={2021}
}

@inproceedings{ssfb-ioidisbomapfansboms-17,
  title={Integration of independence detection into sat-based optimal multi-agent path finding-A novel sat-based optimal MAPF solver},
  author={Surynek, Pavel and {\v{S}}vancara, Ji{\v{r}}{\'\i} and Felner, Ariel and Boyarski, Eli},
  booktitle={International Conference on Agents and Artificial Intelligence},
  volume={2},
  pages={85--95},
  year={2017},
  organization={SciTePress}
}

@inproceedings{lb-pasfcpfwcg-11,
  title={Push and swap: Fast cooperative path-finding with completeness guarantees},
  author={Luna, Ryan and Bekris, Kostas E},
  booktitle={IJCAI},
  pages={294--300},
  year={2011}
}

@article{dtw-paracmapfa-14,
  title={Push and rotate: a complete multi-agent pathfinding algorithm},
  author={De Wilde, Boris and Ter Mors, Adriaan W and Witteveen, Cees},
  journal={Journal of Artificial Intelligence Research},
  volume={51},
  pages={443--492},
  year={2014}
}

@inproceedings{s-anptppfmribcg-09,
  title={A novel approach to path planning for multiple robots in bi-connected graphs},
  author={Surynek, Pavel},
  booktitle={2009 IEEE International Conference on Robotics and Automation},
  pages={3613--3619},
  year={2009},
  organization={IEEE}
}

@inproceedings{khs-aptafnomapf-11,
  title={A polynomial-time algorithm for non-optimal multi-agent pathfinding},
  author={Khorshid, Mokhtar and Holte, Robert and Sturtevant, Nathan},
  booktitle={Proceedings of the International Symposium on Combinatorial Search},
  volume={2},
  number={1},
  pages={76--83},
  year={2011}
}

@inproceedings{slb-mapfwseosap-12,
  title={Multi-agent pathfinding with simultaneous execution of single-agent primitives},
  author={Sajid, Qandeel and Luna, Ryan and Bekris, Kostas},
  booktitle={Proceedings of the International Symposium on Combinatorial Search},
  volume={3},
  number={1},
  pages={88--96},
  year={2012}
}

@inproceedings{bssf-svotcbsftmapfp-14,
  title={Suboptimal variants of the conflict-based search algorithm for the multi-agent pathfinding problem},
  author={Barer, Max and Sharon, Guni and Stern, Roni and Felner, Ariel},
  booktitle={Proceedings of the International Symposium on Combinatorial Search},
  volume={5},
  number={1},
  pages={19--27},
  year={2014}
}

@inproceedings{bfsssbt-icbsfomapf-15,
  title={Improved conflict-based search for optimal multi-agent path finding},
  author={Boyarski, E and Felner, A and Stern, R and Sharon, G and Shimony, E and Bezalel, O and Tolpin, D},
  booktitle={24th International Joint Conference on Artificial Intelligence, IJCAI 2015},
  year={2015}
}

@article{hccbs,
  title={Parallel hierarchical composition conflict-based search for optimal multi-agent pathfinding},
  author={Lee, Hannah and Motes, James and Morales, Marco and Amato, Nancy M},
  journal={IEEE Robotics and Automation Letters},
  volume={6},
  number={4},
  pages={7001--7008},
  year={2021},
  publisher={IEEE}
}

@article{ssgf-tictsfmapf-13,
  title={The increasing cost tree search for optimal multi-agent pathfinding},
  author={Sharon, Guni and Stern, Roni and Goldenberg, Meir and Felner, Ariel},
  journal={Artificial intelligence},
  volume={195},
  pages={470--495},
  year={2013},
  publisher={Elsevier}
  }

@article{wc-sefmrpp-15,
  title={Subdimensional expansion for multirobot path planning},
  author={Wagner, Glenn and Choset, Howie},
  journal={Artificial intelligence},
  volume={219},
  pages={1--24},
  year={2015},
  publisher={Elsevier}
}

@inproceedings{cgssbssh-peawsng-12,
  title={Partial-expansion A* with selective node generation},
  author={Felner, Ariel and Goldenberg, Meir and Sharon, Guni and Stern, Roni and Beja, Tal and Sturtevant, Nathan and Schaeffer, Jonathan and Holte, Robert},
  booktitle={Proceedings of the AAAI Conference on Artificial Intelligence},
  volume={26},
  number={1},
  pages={471--477},
  year={2012}
}

@inproceedings{s-cp-05,
  title={Cooperative pathfinding},
  author={Silver, David},
  booktitle={Proceedings of the aaai conference on artificial intelligence and interactive digital entertainment},
  volume={1},
  number={1},
  pages={117--122},
  year={2005}
}

@inproceedings{s-fostcpp-10,
  title={Finding optimal solutions to cooperative pathfinding problems},
  author={Standley, Trevor},
  booktitle={Proceedings of the AAAI Conference on Artificial Intelligence},
  volume={24},
  number={1},
  pages={173--178},
  year={2010}
}

@inproceedings{pl-ssippfde-11,
  title={Sipp: Safe interval path planning for dynamic environments},
  author={Phillips, Mike and Likhachev, Maxim},
  booktitle={2011 IEEE international conference on robotics and automation},
  pages={5628--5635},
  year={2011},
  organization={IEEE}
}

@article{l-prmfrpp-98,
  title={Probabilistic roadmaps for robot path planning},
  author={Latombe, LEKJ-C},
  journal={Practical motion planning in robotics: current approaches and future challenges},
  pages={33--53},
  year={1998},
  publisher={Citeseer}
}

@article{l-rrtantfpp-98,
  title={Rapidly-exploring random trees: A new tool for path planning},
  journal={Research Report 9811},
  year={1998},
  publisher={Department of Computer Science, Iowa State University}
}

@article{ltw-aafpcfpapo-79,
  title={An algorithm for planning collision-free paths among polyhedral obstacles},
  author={Lozano-P{\'e}rez, Tom{\'a}s and Wesley, Michael A},
  journal={Communications of the ACM},
  volume={22},
  number={10},
  pages={560--570},
  year={1979},
  publisher={ACM New York, NY, USA}
}

@article{ot-vgalva-01,
  title={Visibility graphs and landscape visibility analysis},
  author={O'Sullivan, David and Turner, Alasdair},
  journal={International journal of geographical information science},
  volume={15},
  number={3},
  pages={221--237},
  year={2001},
  publisher={Taylor \& Francis}
}

@incollection{f-vdadt-17,
  title={Voronoi diagrams and Delaunay triangulations},
  author={Fortune, Steven},
  booktitle={Handbook of discrete and computational geometry},
  pages={705--721},
  year={2017},
  publisher={Chapman and Hall/CRC}
}

@inproceedings{was-aprmpwsotmaotfs-99,
  title={MAPRM: A probabilistic roadmap planner with sampling on the medial axis of the free space},
  author={Wilmarth, Steven A and Amato, Nancy M and Stiller, Peter F},
  booktitle={Proceedings 1999 IEEE international conference on robotics and automation (Cat. No. 99CH36288C)},
  volume={2},
  pages={1024--1031},
  year={1999},
  organization={IEEE}
}

@article{a-vdasoafgds-91,
  title={Voronoi diagrams—a survey of a fundamental geometric data structure},
  author={Aurenhammer, Franz},
  journal={ACM Computing Surveys (CSUR)},
  volume={23},
  number={3},
  pages={345--405},
  year={1991},
  publisher={ACM New York, NY, USA}
}

@article{b-afafbc-01,
  title={A faster algorithm for betweenness centrality},
  author={Brandes, Ulrik},
  journal={Journal of mathematical sociology},
  volume={25},
  number={2},
  pages={163--177},
  year={2001},
  publisher={Taylor \& Francis}
}

@inproceedings{l-mapffla-19,
  title={Multi-agent path finding for large agents},
  author={Li, Jiaoyang and Surynek, Pavel and Felner, Ariel and Ma, Hang and Kumar, TK Satish and Koenig, Sven},
  booktitle={Proceedings of the AAAI Conference on Artificial Intelligence},
  volume={33},
  number={01},
  pages={7627--7634},
  year={2019}
}

@inproceedings{o-lsbafqmapf-23,
  title={Lacam: Search-based algorithm for quick multi-agent pathfinding},
  author={Okumura, Keisuke},
  booktitle={Proceedings of the AAAI Conference on Artificial Intelligence},
  volume={37},
  number={10},
  pages={11655--11662},
  year={2023}
}

@inproceedings{bfhsclk-idcbs-20,
  title={Iterative-deepening conflict-based search},
  author={Boyarski, Eli and Felner, Ariel and Harabor, Daniel and Stuckey, Peter J and Cohen, Liron and Li, Jiaoyang and Koenig, Sven},
  booktitle={International Joint Conference on Artificial Intelligence-Pacific Rim International Conference on Artificial Intelligence 2020},
  pages={4084--4090},
  year={2020},
  organization={Association for the Advancement of Artificial Intelligence (AAAI)}
}

@inproceedings{clbmckk-ahtcbsfmapf-18,
  title={Adding heuristics to conflict-based search for multi-agent path finding},
  author={Felner, Ariel and Li, Jiaoyang and Boyarski, Eli and Ma, Hang and Cohen, Liron and Kumar, TK Satish and Koenig, Sven},
  booktitle={Proceedings of the International Conference on Automated Planning and Scheduling},
  volume={28},
  pages={83--87},
  year={2018}
}

@inproceedings{lrk-eecbsabssfmapf-21,
  title={Eecbs: A bounded-suboptimal search for multi-agent path finding},
  author={Li, Jiaoyang and Ruml, Wheeler and Koenig, Sven},
  booktitle={Proceedings of the AAAI conference on artificial intelligence},
  volume={35},
  number={14},
  pages={12353--12362},
  year={2021}
}

@inproceedings{lfbmk-ihfmapfwcbs-19,
  title={Improved Heuristics for Multi-Agent Path Finding with Conflict-Based Search.},
  author={Li, Jiaoyang and Felner, Ariel and Boyarski, Eli and Ma, Hang and Koenig, Sven},
  booktitle={IJCAI},
  volume={2019},
  pages={442--449},
  year={2019}
}

@inproceedings{lghsmk-ntfpsbimapf-20,
  title={New techniques for pairwise symmetry breaking in multi-agent path finding},
  author={Li, Jiaoyang and Gange, Graeme and Harabor, Daniel and Stuckey, Peter J and Ma, Hang and Koenig, Sven},
  booktitle={Proceedings of the International Conference on Automated Planning and Scheduling},
  volume={30},
  pages={193--201},
  year={2020}
}

@inproceedings{lhsfmk-dsfmapfwcbs-19,
  title={Disjoint splitting for multi-agent path finding with conflict-based search},
  author={Li, Jiaoyang and Harabor, Daniel and Stuckey, Peter J and Felner, Ariel and Ma, Hang and Koenig, Sven},
  booktitle={Proceedings of the international conference on automated planning and scheduling},
  volume={29},
  pages={279--283},
  year={2019}
}

@inproceedings{lhsmk-sbcfgbmapf-19,
  title={Symmetry-breaking constraints for grid-based multi-agent path finding},
  author={Li, Jiaoyang and Harabor, Daniel and Stuckey, Peter J and Ma, Hang and Koenig, Sven},
  booktitle={Proceedings of the AAAI conference on artificial intelligence},
  volume={33},
  number={01},
  pages={6087--6095},
  year={2019}
}

@inproceedings{ssfs-macbsfomapf-12,
  title={Meta-agent conflict-based search for optimal multi-agent path finding},
  author={Sharon, Guni and Stern, Roni and Felner, Ariel and Sturtevant, Nathan},
  booktitle={Proceedings of the International Symposium on Combinatorial Search},
  volume={3},
  number={1},
  pages={97--104},
  year={2012}
}

@article{aysas-mapfwct-22,
  title={Multi-agent pathfinding with continuous time},
  author={Andreychuk, Anton and Yakovlev, Konstantin and Surynek, Pavel and Atzmon, Dor and Stern, Roni},
  journal={Artificial Intelligence},
  volume={305},
  pages={103662},
  year={2022},
  publisher={Elsevier}
}

@inproceedings{svsab-omapf-19,
  title={Online multi-agent pathfinding},
  author={{\v{S}}vancara, Ji{\v{r}}{\'\i} and Vlk, Marek and Stern, Roni and Atzmon, Dor and Bart{\'a}k, Roman},
  booktitle={Proceedings of the AAAI conference on artificial intelligence},
  volume={33},
  number={01},
  pages={7732--7739},
  year={2019}
}

@article{jzzjyh-cbswdlafrppiude-23,
  title={Conflict-based search with D* lite algorithm for robot path planning in unknown dynamic environments},
  author={Jin, Jianzhi and Zhang, Yin and Zhou, Zhuping and Jin, Mengyuan and Yang, Xiaolian and Hu, Fang},
  journal={Computers and Electrical Engineering},
  volume={105},
  pages={108473},
  year={2023},
  publisher={Elsevier}
}

@inproceedings{a-mapfwkcvcbs-20,
  title={Multi-agent path finding with kinematic constraints via conflict based search},
  author={Andreychuk, Anton},
  booktitle={Artificial Intelligence: 18th Russian Conference, RCAI 2020, Moscow, Russia, October 10--16, 2020, Proceedings 18},
  pages={29--45},
  year={2020},
  organization={Springer}
}

@inproceedings{boyarski2015don,
  title={Don't split, try to work it out: Bypassing conflicts in multi-agent pathfinding},
  author={Boyarski, Eli and Felner, Ariel and Sharon, Guni and Stern, Roni},
  booktitle={Proceedings of the International Conference on Automated Planning and Scheduling},
  volume={25},
  pages={47--51},
  year={2015}
}

@inproceedings{walker2021conflict,
  title={Conflict-based increasing cost search},
  author={Walker, Thayne T and Sturtevant, Nathan R and Felner, Ariel and Zhang, Han and Li, Jiaoyang and Kumar, TK Satish},
  booktitle={Proceedings of the International Conference on Automated Planning and Scheduling},
  volume={31},
  pages={385--395},
  year={2021}
}

@article{andreychuk2022multi,
  title={Multi-agent pathfinding with continuous time},
  author={Andreychuk, Anton and Yakovlev, Konstantin and Surynek, Pavel and Atzmon, Dor and Stern, Roni},
  journal={Artificial Intelligence},
  volume={305},
  pages={103662},
  year={2022},
  publisher={Elsevier}
}

@inproceedings{gange2019lazy,
  title={Lazy CBS: implicit conflict-based search using lazy clause generation},
  author={Gange, Graeme and Harabor, Daniel and Stuckey, Peter J},
  booktitle={Proceedings of the international conference on automated planning and scheduling},
  volume={29},
  pages={155--162},
  year={2019}
}

@article{zhang2022multi,
  title={Multi-agent path finding with mutex propagation},
  author={Zhang, Han and Li, Jiaoyang and Surynek, Pavel and Kumar, TK Satish and Koenig, Sven},
  journal={Artificial Intelligence},
  volume={311},
  pages={103766},
  year={2022},
  publisher={Elsevier}
}

@inproceedings{chan2023greedy,
  title={Greedy priority-based search for suboptimal multi-agent path finding},
  author={Chan, Shao-Hung and Stern, Roni and Felner, Ariel and Koenig, Sven},
  booktitle={Proceedings of the International Symposium on Combinatorial Search},
  volume={16},
  number={1},
  pages={11--19},
  year={2023}
}

@inproceedings{xu2022multi,
  title={Multi-goal multi-agent pickup and delivery},
  author={Xu, Qinghong and Li, Jiaoyang and Koenig, Sven and Ma, Hang},
  booktitle={2022 IEEE/RSJ International Conference on Intelligent Robots and Systems (IROS)},
  pages={9964--9971},
  year={2022},
  organization={IEEE}
}

@article{zhang2024d,
  title={D-PBS: Dueling priority-based search for multiple nonholonomic robots motion planning in congested environments},
  author={Zhang, Xiaotong and Xiong, Gang and Wang, Yuanjing and Teng, Siyu and Chen, Long},
  journal={IEEE Robotics and Automation Letters},
  year={2024},
  publisher={IEEE}
}

@inproceedings{chen2021scalable,
  title={Scalable and safe multi-agent motion planning with nonlinear dynamics and bounded disturbances},
  author={Chen, Jingkai and Li, Jiaoyang and Fan, Chuchu and Williams, Brian C},
  booktitle={Proceedings of the AAAI conference on artificial intelligence},
  volume={35},
  number={13},
  pages={11237--11245},
  year={2021}
}

@article{solis2021representation,
  title={Representation-optimal multi-robot motion planning using conflict-based search},
  author={Solis, Irving and Motes, James and Sandstr{\"o}m, Read and Amato, Nancy M},
  journal={IEEE Robotics and Automation Letters},
  volume={6},
  number={3},
  pages={4608--4615},
  year={2021},
  publisher={IEEE}
}

@inproceedings{cohen2010search,
  title={Search-based planning for manipulation with motion primitives},
  author={Cohen, Benjamin J and Chitta, Sachin and Likhachev, Maxim},
  booktitle={2010 IEEE international conference on robotics and automation},
  pages={2902--2908},
  year={2010},
  organization={IEEE}
}

@inproceedings{li2021anytime, 
  title={Anytime multi-agent path finding via large neighborhood search},
  author={Li, Jiaoyang and Chen, Zhe and Harabor, Daniel and Stuckey, Peter J and Koenig, Sven},
  booktitle={International Joint Conference on Artificial Intelligence 2021},
  pages={4127--4135},
  year={2021},
  organization={Association for the Advancement of Artificial Intelligence (AAAI)}
}

@inproceedings{li2022mapf,
  title={MAPF-LNS2: Fast repairing for multi-agent path finding via large neighborhood search},
  author={Li, Jiaoyang and Chen, Zhe and Harabor, Daniel and Stuckey, Peter J and Koenig, Sven},
  booktitle={Proceedings of the AAAI Conference on Artificial Intelligence},
  volume={36},
  number={9},
  pages={10256--10265},
  year={2022}
}

@inproceedings{wang2025lns2+,
  title={LNS2+ RL: Combining multi-agent reinforcement learning with large neighborhood search in multi-agent path finding},
  author={Wang, Yutong and Duhan, Tanishq and Li, Jiaoyang and Sartoretti, Guillaume},
  booktitle={Proceedings of the AAAI Conference on Artificial Intelligence},
  volume={39},
  number={22},
  pages={23343--23350},
  year={2025}
}
\begin{IEEEbiography}[{\includegraphics[width=1in,height=1.25in,clip]{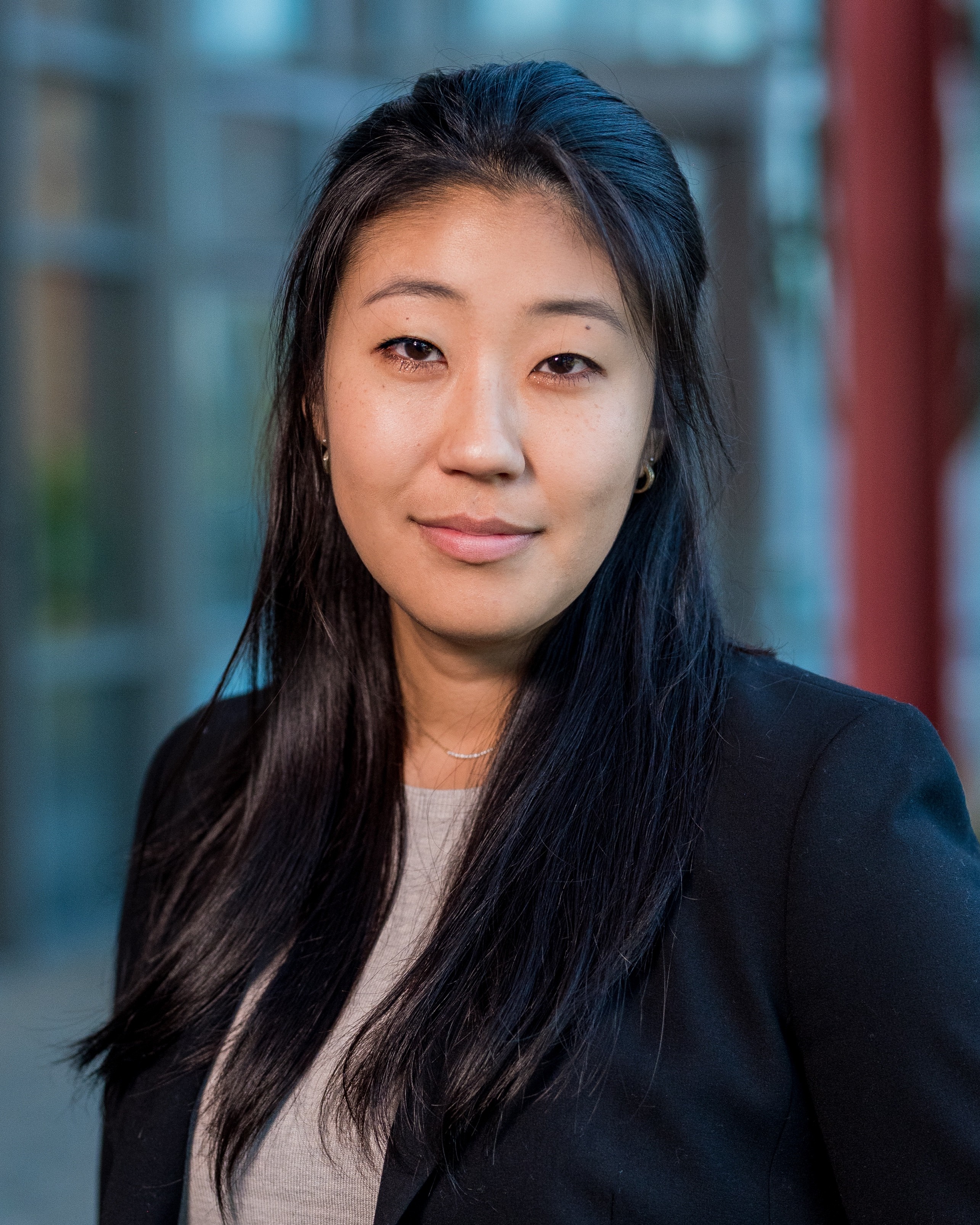}}]{Hannah Lee}
is a Software Engineer at Anduril Industries, where she works on autonomy and motion planning. She received her Ph.D. in Computer Science from the University of Illinois Urbana Champaign (UIUC) and her B.S. in Computer Science from the Colorado School of Mines. Her research interests include multi-agent pathfinding, multi-robot planning, and artificial intelligence.
\end{IEEEbiography}

\begin{IEEEbiography}[{\includegraphics[width=1in,height=1.25in,clip]{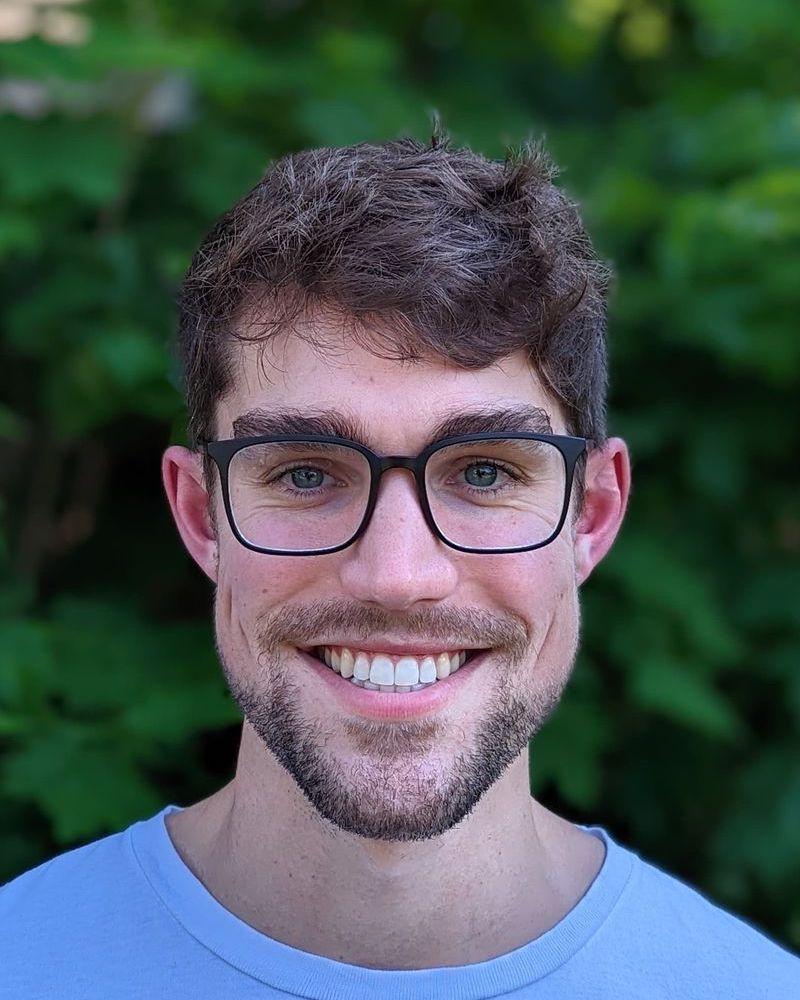}}]{James D. Motes}
is a Postdoctoral Researcher in the Siebel School of Computing and Data Science at the University of Illinois at Urbana Champaign (UIUC) where he also completed his Ph.D. in Computer Science.
He received both a B.S. in Computer Engineering and a M.S. degree in Computer Science from Texas A\&M University.
His research interests include task and motion planning, multi-robot systems, and autonomous factories.
\end{IEEEbiography}

\begin{IEEEbiography}[{\includegraphics[width=1in,height=1.25in,clip,keepaspectratio]{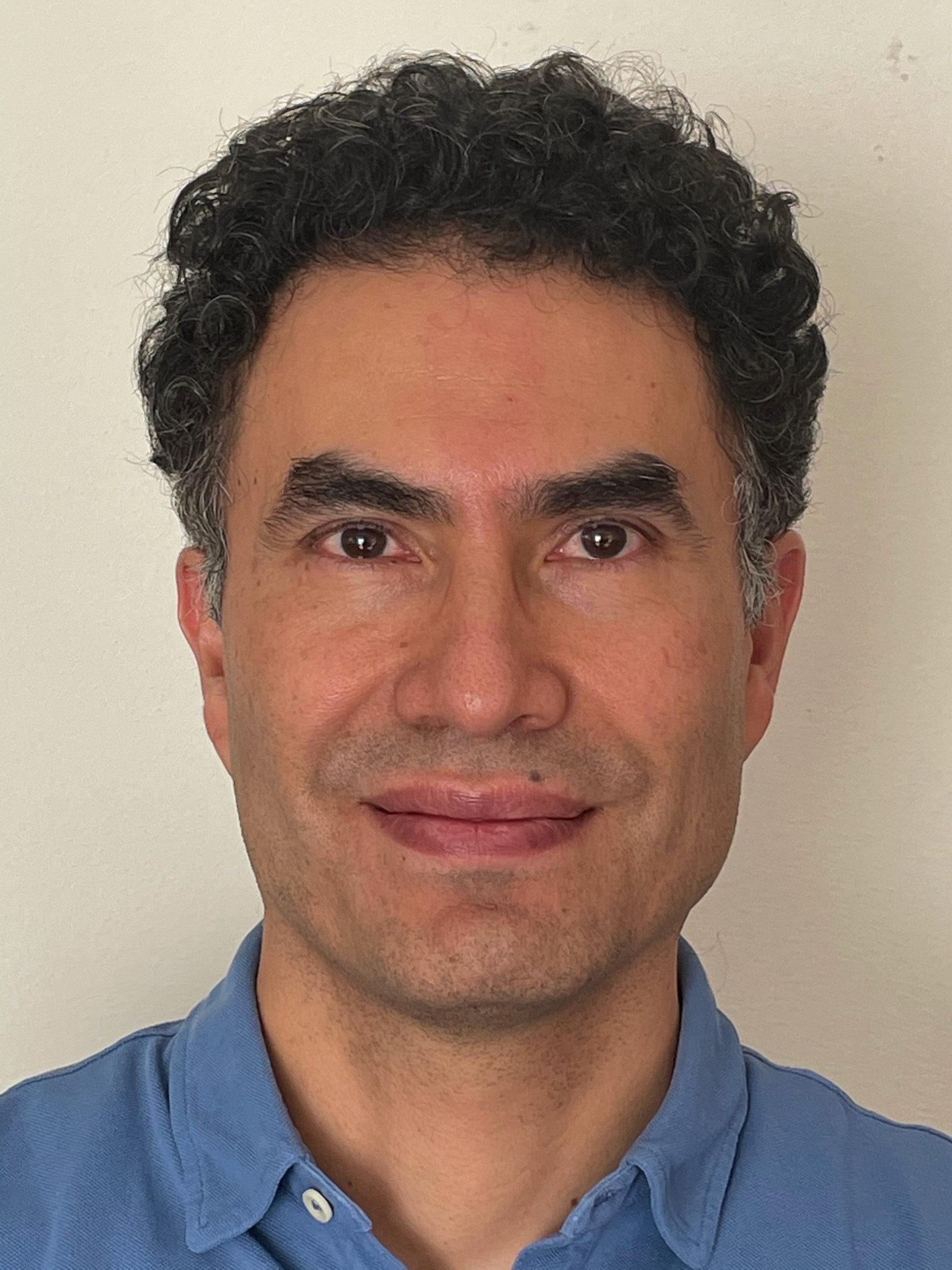}}]{Marco Morales Aguirre}
is an Associate Professor in the Department of Computer Science at Instituto Tecnológico Autónomo de México (ITAM) and a Teaching Associate Professor in the Siebel School of Computing and Data Science at the University of Illinois Urbana Champaign. He is a National Researcher (Level II) within the National System of Researchers of Mexico and a member of the Mexican Academy of Computing. He has also been a Visiting Professor at Texas A\&M University and a Lecturer at Universidad Nacional Autónoma de México (UNAM). He holds a Ph.D. in Computer Science from Texas A\&M University, a M.S. in Electrical Engineering and a B.S. in Computer Engineering from UNAM. His research interests are in motion planning and control for autonomous robots, artificial intelligence, machine learning, and computational geometry.
\end{IEEEbiography}

% \vskip -2.5\baselineskip plus -1fil

\begin{IEEEbiography}[{\includegraphics[width=1in,height=1.25in,clip,keepaspectratio]{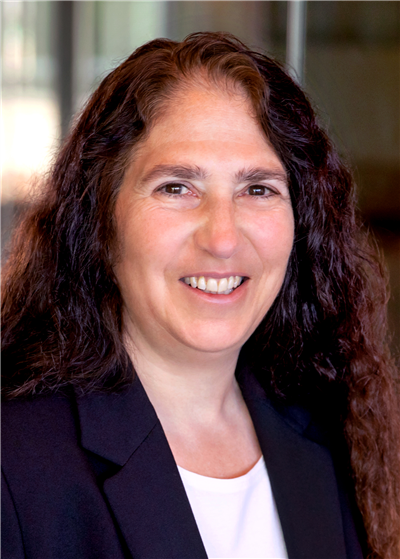}}]{Nancy M. Amato}
(F'10) is Director of the Siebel School of Computing and Data Science and Abel Bliss Professor of Engineering at the University of Illinois at Urbana Champaign.
She received undergraduate degrees in Mathematical Sciences and Economics
from Stanford, and M.S. and Ph.D. degrees in Computer Science from UC Berkeley
and the University of Illinois, respectively.
Before returning to her alma mater in 2019, she was Unocal Professor and
Regents Professor at Texas A\&M University and Senior Director of Engineering Honors Programs.
Amato is known for algorithmic contributions to robotics task and motion planning,
computational biology and geometry, and parallel and distributed computing.
Her honors include the inaugural Robotics Medal in 2023,
the 2019 IEEE RAS Saridis Leadership Award in Robotics and Automation,
the inaugural 2014 NCWIT Harrold/Notkin Research and Graduate Mentoring Award,
and the 2014 CRA Haberman Award for contributions to increasing diversity in computing.
She is a Fellow of the AAAI, AAAS, ACM, and IEEE.
\end{IEEEbiography}

\vfill

\end{document}